\RequirePackage{fix-cm}
\documentclass[twocolumn]{svjour3}          % twocolumn
\smartqed  % flush right qed marks, e.g. at end of proof
\usepackage{subcaption}

\usepackage{graphicx}
\usepackage{pgfplots} 
\pgfplotsset{compat=1.11,
    /pgfplots/ybar legend/.style={
    /pgfplots/legend image code/.code={%
       \draw[##1,/tikz/.cd,yshift=-0.25em]
        (0cm,0cm) rectangle (3pt,0.8em);},
   },
}
\usepackage{tikz}
\usetikzlibrary{shapes,arrows}

\usepackage{amsmath}

\usepackage{booktabs}

\usepackage{hyperref}

\usepackage{xurl}
\usepackage{breakcites}

\usepackage{float}
\floatstyle{plaintop}
\restylefloat{table}

\usepackage{natbib} 

\usepackage[misc]{ifsym}

\usepackage{xspace}
\usepackage{fixfoot}
\DeclareFixedFootnote*{\TechReport}{\textcolor{black}{\url{https://arxiv.org/pdf/abcd.efgh.pdf}}}

\definecolor{BLACK}{rgb}{0,0,0}
\definecolor{WHITE}{rgb}{1,1,1}
\definecolor{RED}{rgb}{1,0.0,0.0}
\definecolor{BLUE2}{rgb}{0.0,0.3,0.6}
\definecolor{DARKSILVER}{rgb}{0.5,0.5,0.5}
\definecolor{CYAN}{rgb}{0.0,1.0,1.0}
\definecolor{GREEN2}{rgb}{0.2,1.0,0.0}
\definecolor{YELLOW}{rgb}{1.0,0.88,0.21}
\definecolor{DARKRED}{rgb}{0.5,0.0,0.13}
\definecolor{LIGHTRED}{rgb}{0.8,0.0,0.0}
\definecolor{LIGHTPURPLE}{rgb}{0.75,0.58,0.89}
\definecolor{PURPLE}{rgb}{0.54,0.17,0.89}
\definecolor{CYAN}{rgb}{0.3,0.91,0.87}
\definecolor{BLUEGREEN}{rgb}{0.11,0.67,0.84}
\definecolor{DARKGREEN}{rgb}{0.0,0.5,0.0}
\definecolor{LIGHTGREEN}{rgb}{0.4,1.0,0.0}
\definecolor{SILVER}{rgb}{0.75,0.75,0.75}
\definecolor{DARKSILVER}{rgb}{0.5,0.5,0.5}
\definecolor{BLUEBROWN}{rgb}{0.52,0.4,0.27}

\newcommand*{\affaddr}[1]{#1} 
\newcommand*{\affmark}[1][*]{\textsuperscript{#1}}

\begin{document}\sloppy

\title{Recent Ice Trends in Swiss Mountain Lakes: 20-year Analysis of MODIS Imagery}
\subtitle{}

%\titlerunning{Short form of title}        % if too long for running head

\author{Manu Tom\protect\affmark[$\,$1,2,3]$\!\!$ \and $\!$Tianyu Wu\protect\affmark[$\,$1]$\!\!$ \and $\!$Emmanuel Baltsavias\protect\affmark[$\,$1]$\!\!$ \and $\!$Konrad Schindler\protect\affmark[$\,$1]}

%\author{Manu Tom \and Tianyu Wu \and Emmanuel Baltsavias \and Konrad Schindler}

\authorrunning{Tom et al.} % if too long for running head

\institute{Manu Tom~(\Letter)\at
          \email{manu.tom@eawag.ch} %  \\
            %\emph{Present address:} of F. Author  %  if needed
           \and
           Tianyu Wu \at
           \email{wuti@student.ethz.ch}
          \and
           Emmanuel Baltsavias \at
           \email{emmanuel.baltsavias@geod.baug.ethz.ch}
           \and
           Konrad Schindler \at
           \email{schindler@ethz.ch}\\
           \\
           \affaddr{\affmark[1]Photogrammetry and Remote Sensing Group,\\ ETH Zurich, 8093 Zurich, Switzerland}\\
           \\
           \affaddr{\affmark[2]Glaciology and Geomorphodynamics Group,\\ University of Zurich, 8057 Zurich, Switzerland
          }\\
          \\
          \affaddr{\affmark[3]Remote Sensing Group, Swiss Federal Institute of Aquatic\\ Science and Technology, 8600 Dübendorf, Switzerland}\\
          \\
          \textbf{This version of the article has been accepted for publication, after peer review but is not the Version of Record and does not reflect post-acceptance improvements, or any corrections. The Version of Record is available online at: http://dx.doi.org/10.1007/s41064-022-00215-x}
}

\date{Received: date / Accepted: date}
% The correct dates will be entered by the editor

\maketitle

\begin{abstract}
Depleting lake ice is a climate change indicator, just like sea-level rise or glacial retreat. Monitoring Lake Ice Phenology (LIP) is useful because long-term freezing and thawing patterns serve as sentinels to understand regional and global climate change. We report a study for the Oberengadin region of Switzerland, where several small- and medium-sized mountain lakes are located. We observe the LIP events, such as freeze-up, break-up and ice cover duration, across two decades (2000-2020) from optical satellite images. We analyse the time series of MODIS imagery by estimating spatially resolved maps of lake ice for these Alpine lakes with supervised machine learning. To train the classifier we rely on reference data annotated manually based on webcam images. From the ice maps, we derive long-term LIP trends. Since the webcam data are only available for two winters, we cross-check our results against the operational MODIS and VIIRS snow products. We find a change in complete freeze duration of $-0.76$ and $-0.89$ days per annum for lakes Sils and Silvaplana, respectively. Furthermore, we observe plausible correlations of the LIP trends with climate data measured at nearby meteorological stations. We notice that mean winter air temperature has a negative correlation with the freeze duration and break-up events and a positive correlation with the freeze-up events. Additionally, we observe a strong negative correlation of sunshine during the winter months with the freeze duration and break-up events.\\ \\
Abnehmende Vereisung von Seen ist, wie der Anstieg des Meeresspiegels oder das Abschmelzen von Gletschern, ein Indikator für den Klimawandel. Die Beobachtung von
ph\"anologischen Ereignissen wie Beginn und Ende der Vereisung ist n\"utzlich, weil diese Anhaltspunkte für die Beschreibung des regionalen und globalen Klimawandels bieten. Dieser Beitrag berichtet \"uber eine Fallstudie zu kleinen bis mittelgrossen Seen in der Region Oberengadin (Schweiz). Beobachtet werden Beginn und Ende der Vereisung sowie die Dauer der Eisbedeckung über zwei Jahrzehnte (2000-2020), anhand von optischen Satellitenbildern. Auf Basis von MODIS-Bildfolgen (sowie VIIRS-Bildfolgen, soweit verf\"ugbar) wurden, mit Hilfe von maschinellem Lernen, explizite Karten des Seeeises erstellt. Als Trainingsdaten für den Klassifikator dienen h\"andische Annotationen auf der Grundlage von Webcam-Bildern. Aus den Eiskarten wurden Trends f\"ur die Seevereisung abgeleitet. Da Webcam-Daten nur für zwei Winter verf\"ugbar sind wurden die Ergebnisse auch mit operationellen MODIS- und VIIRS-Schneeprodukten verglichen. Es zeigt sich eine Verk\"urzung der Vereisungsdauer um $0.76$ bzw. $0.89$ Tage pro Jahr für den Silsersee und den Silvaplanersee. Weiters beobachten wir plausible Korrelationen der Trends mit Klimadaten von nahe gelegenen meteorologischen Stationen. Konkret korreliert die mittlere Lufttemperatur der Wintermonate stark mit Beginn, Ende und Dauer der Vereisung. Ebenso gibt es eine starke Korrelation zwischen der Sonnenscheindauer w\"ahrend der Wintermonate und dem Ende sowie der Dauer der Vereisung.
% \PACS{PACS code1 \and PACS code2 \and more}
%\subclass{MSC code1 \and MSC code2 \and more}
\end{abstract}

%{\obeylines}
\vspace{-0.75em}
\section{Introduction}
\vspace{-0.5em}
Lake ice cover is part of the essential climate variable: \emph{lakes} (\url{https://gcos.wmo.int/en/essential-climate-variables/lakes/}).
Many studies have reported about the response of Lake Ice Phenology (LIP) to climate variations \citep{Brown_and_Duguay_2010,Duguay2006journal,Howell_2009,Kang_2012,Sharma2019_Journal_Nature_ClimateChange,SurduDuguayBrown_2014}. Local weather patterns and lake ice formation processes are inter-connected~\citep{Brown_and_Duguay_2010}. Hence, monitoring the long-term LIP trends can provide integral cues on the local and global climate. Increasing temperatures cause decreasing trends in the lake ice formation process. Air temperature in the vicinity of a lake affects the ice formation process within the lake and vice versa. Moreover, there are potential positive feedbacks, as frozen lakes have a higher albedo (especially when covered with snow), and thus lower absorption and evaporation~\citep{Slater_2021,Wang_2018}.
In addition to its contribution to climate studies, lake ice monitoring is also useful to organise safe transportation, especially in lakes that freeze partially, to conserve freshwater ecosystems, to trigger warnings against ice shoves caused by wind during the break-up period, and for winter tourism~\citep{Hampton2017,Hirose2008,Knoll2019,Mullan2017}. 
\par
In this study, we monitor the spatio-temporal extent\footnote{We measure ice cover, but not ice thickness.} of ice on lakes of the Oberengadin region in the Swiss Alps (which reliably freeze every winter) daily over 20 winters.
From those time series, we derive the dates of the following important LIP events: Freeze-Up Start (FUS), Freeze-Up End (FUE), Break-Up Start (BUS) and Break-Up End (BUE). Using these four dates, we also estimate the Complete Freeze Duration (CFD) and Ice Coverage Duration (ICD).
For lake ice monitoring, the Global Climate Observing System (GCOS) office requirement are daily observations, and an accuracy of $\pm2$ days for the \emph{ice-on/-off} dates~(\url{https://gcos.wmo.int/en/essential-climate-variables/lakes/ecv-requirements}).
\par
In this work, we use only image data from optical satellites\footnote{We have previously also used webcams~\citep{Tom2020_lakeice_RS_journal,muyan_lakeice_2018} and Sentinel-1 Synthetic Aperture Radar~\citep[SAR]{tom_aguilar_2020} for lake ice monitoring.}, and provide a direct, data-driven observation not influenced by model assumptions about the ice formation process. 
We see satellite imagery as an independent information source and consider image analysis complementary to other methods of lake ice modelling. MODIS and VIIRS have several advantages such as wide area coverage, good spectral and fine temporal resolution (daily), and free availability. Additionally, as opposed to other optical satellites such as Landsat-8, Sentinel-2 and the like, MODIS and VIIRS offer the best spatio-temporal resolution trade-off for the application of single-sensor lake ice monitoring, even though the spatial resolution is moderate (250-1000m Ground Sampling Distance, GSD). An important asset is the availability of long time series, e.g., MODIS data is available for the entire period since 2000, contrary to other sensor data like airborne or terrestrial photography, webcams etc.
We use the linear Support Vector Machine~\citep[SVM]{SVM} classifier to perform semantic segmentation and derive the LIP events from the resulting time series by fitting a piece-wise linear model per winter. Additionally, we perform a 4-fold cross-validation experiment and assess the transferability of the learned model across space and time.
\par
To our knowledge, the only operational lake ice product is the Climate Change Initiative Lake Ice Cover~\citep{CCI_lakes}, however, our target lakes are not included among the 250 lakes it covers. A second product, Copernicus Lake Ice Extent ~(LIE, \url{https://land.copernicus.eu/global/products/lie}), is still in the pre-operational stage due to accuracy issues, and coverage only starts in 2017. Though not designed for lake ice, the MODIS Snow Product~\citep[MSP]{snowmap} and VIIRS Snow Product~(VSP, \url{https://nsidc.org/sites/nsidc.org/files/technical-references/VIIRS-snow-products-user-guide-final.pdf}) are also reasonable proxies since lakes in the Alps are typically snow-covered for most of the frozen period. In \cite{Tom2020_lakeice_RS_journal}, we have reported a comparison of specifications of the operational lake ice/snow products. We cross-check our results with these two snow products, see Section \ref{section:modis_20winters}. 
\par
For Swiss lakes, a previous study~\citep{franssen_scherrer2008} has verified that the lake ice formation is strongly correlated with the surrounding air temperature. The authors deducted an empirical relationship between the sum of negative degree days (also called Accumulated Freezing Degree Days, AFDD) and the lake ice formation process, and modelled the probability of ice cover via binomial logistic regression. For that study, temperature data from 1901-2006 was gathered to study eleven lakes in the lower-lying Swiss plateau. However, none of the mountain lakes we target were included.
\par
The aims of the proposed study are twofold. First, it supplements the existing databases about lake ice on mountain lakes, and further corroborates the accuracy of supervised machine learning for detecting LIP events on small lakes, using only low-resolution imagery. Second, it examines the correlation between LIP and climate data for small lakes in a topographically and climatically complex Alpine environment.
\vspace{-2em}
\subsection{LIP trend analysis studies with MODIS and VIIRS}
\vspace{-0.25em}
The LIP trends of several lakes with different geographical conditions have been studied and reported in the literature. Though most of them used information from various ice databases~(e.g., NSIDC), some studies directly derived the trends from radar and optical satellite data. In the following, we focus on studies that, like ours, analyse MODIS and/or VIIRS optical satellite imagery to examine trends over multiple winters.
\par
\cite{Latifovic2007} used the AVHRR historical data record in addition to in-situ measurements to perform long-term (1950-2004) trend analysis of Canadian lakes with areas $>100~km^{2}$, via an automated profile feature extraction procedure. They confirmed later freeze-up ($0.12$ d/a) and earlier break-up ($-0.18$ d/a) for the majority of lakes that were analysed and suggested that their procedure to extract the LIP events is not sensor-specific and could be applied to other satellite data, too. \cite{Murfitt_Brown_2017} also used MODIS data to extract lake ice trends (2001-2014) for the Canadian states of Ontario and Manitoba, and found regionally varying trends.
\par
\cite{Zhang2019_Journal_MODIS} proposed to threshold the red reflectance band of MODIS (threshold values determined with the help of Landsat images) to monitor ice on lakes in Maine (USA), with areas ranging from 0.13 to $305~km^2$. Though they analysed data from 296 lakes over 19 years (2000-2018), significant break-up and freeze-up trends were detected only for $3\%$, respectively $2.1\%$ of the lakes. The same threshold-based approach, with additional post-processing filters, was applied to a new LIP database covering 4241 Alaskan lakes~\citep{Zhang_LIP_Alaska_2021} with areas $>1~km^{2}$, over the period 2000-2019. The estimated significant LIP trends are: later freeze-up ($0.29$ d/a) and earlier break-up ($-0.55$ d/a) for 289 and 440 lakes, respectively; and earlier freeze-up ($-0.33$ d/a) and later break-up ($0.75$ d/a) for only 11 and 4 lakes, respectively.
\par
\cite{smejkalova2016_Journal_iceoff} extracted the LIP trends (2000–2013) for 13,300 Arctic lakes (area $>$1~$km^{2}$) using MODIS imagery, and observed a trend towards earlier break-up. They reported a mean shift in BUS in the range: $-0.10$ d/a (Northern Europe) to $-1.05$ d/a (central Siberia), and BUE in the range: $-0.14$ d/a to $-0.72$ d/a.
\cite{lakeice_MODIS_Tibet2013} studied the LIP trends of 59 lakes (area $>100~km^{2}$) on the Tibetan Plateau from 2001 to 2010 using MODIS data. However, the estimated LIP trends varied across the target lakes and it was concluded that the 10-year time span is too short to draw a firm conclusion about LIP trends.
\cite{Gou2015} analysed the ice formation trends (2000-2013) in lake Nam Co (Tibet, area $1920~km^{2}$) using MODIS and in-situ data and found strong correlations with air temperature and wind speed patterns. This study found that high wind speeds during winter time could potentially expedite the freeze-up process. Additionally, this work reported a significant reduction in the total freeze duration. \cite{Gou2017} later analysed Nam Co for the period 2000 till 2015 using multiple MODIS products and reported delayed FUS ($0.58$ d/a) as well as BUS ($0.09$ d/a), and reduced ice duration ($-0.49$ d/a) trends. Another study~\citep{Yao2016} also noted an increasingly shorter freeze duration during the period 2000-2011 when investigating the lakes in the Hoh Xil region (Tibet, 22 lakes with area $>100~km^{2}$), using MODIS, Landsat TM/ETM+, and meteorological data. In addition, that work estimated later freeze-up and earlier break-up trends. They reported that the FUS, FUE, BUS, BUE, CFD and ICD shifted on average by $0.73$, $0.34$, $-1.66$, $-0.81$, $-1.91$, $-2.21$ d/a respectively. \cite{Cai_MODIS_Tibet2019} also analysed 58 lakes (area $>41~km^{2}$) located on the Tibetan Plateau during the period from 2001 till 2017 using both Terra and Aqua MODIS imagery. For 47 lakes, a later FUS was noticed ($0.55$ d/a) while for the remaining 11 lakes an earlier FUS was observed ($-0.44$ d/a). For 50\% of the target lakes, an earlier BUE ($-0.69$ d/a) was noted, however, for the other half a later BUE ($0.39$ d/a) was observed. Additionally, they reported a reduced ice cover duration for 40 lakes ($-0.8$ d/a), while for 18 lakes an increase was noted ($1.11$ d/a).
\par
\cite{Yang2019} used MODIS to estimate the LIP trends for 8 large lakes ($106$ to $3461~km^{2}$) in Northeastern China from 2003 to 2016. Later FUS ($0.65$ d/a), earlier BUE ($-0.19$ d/a) and shorter freeze duration ($-0.84$ d/a) trends were noticed. \cite{Qinghai2020} used AVHRR, MODIS, and Landsat data to extract the LIP of Qinghai lake (China, area of $4294~km^{2}$) for the period 1980-2018. They estimated a shift of $0.16$, $0.19$, $-0.36$, and $-0.42$ d/a for FUS, FUE, BUS and BUE respectively, also pointing towards progressively later freeze-up and earlier break-up. Additionally, they computed the decreasing patterns in ICD ($-0.58$ d/a) and CFD ($-0.52$ d/a). That study also identified correlations between the LIP and climate indicators like the AFDD, wind speed, precipitation, etc.\ during the winter season. \cite{Cai_MODIS_China2020} used a threshold-based method to extract LIP trends from MODIS snow product for 23 lakes (2001-2018, Xinjiang Uygur Region, area: $11$ to $1004~km^{2}$) in China. They found that the ICD decreased ($-1.08$ d/a) in 16 out of the 23 lakes and increased ($1.18$ d/a) for the rest. In addition, they reported later freeze-up ($0.52$ d/a) and earlier break-up ($-0.51$ d/a) in 17 and 18 lakes, respectively. Additionally, they found that the freeze-up events are more affected by lake-specific factors such as area and mineralisation; while climatic factors like Lake Surface Water Temperature (LSWT) have more influence on the break-up events. That work also emphasised that LSWT has a stronger influence on the LIP events than the air temperature.
\par
In this work, we deal with small- and medium-sized lakes, whereas most trend studies that used MODIS~\citep{Gou2015,Qi2019,Qinghai2020,Yao2016} focused on large lakes. Notable exceptions are a study by \cite{smejkalova2016_Journal_iceoff}, concerning only break-up trends; and by \cite{Zhang2019_Journal_MODIS}, who reported results for lakes as small as $0.13~km^{2}$, but with a limited accuracy of 5-8 days mean absolute error for the ice-on/-off dates, and with no significant trend. In \cite{Zhang_LIP_Alaska_2021}, the same team processed lakes down to $1~km^{2}$ with an accuracy of 5-11 days. It is interesting to note that, unlike the present study, both these works \citep{Zhang2019_Journal_MODIS, Zhang_LIP_Alaska_2021} did not perform any correction for absolute geo-location error when handling small lakes of only a few, often mixed, MODIS pixels. They also found it necessary to tune the thresholds separately for each lake and even use different thresholds for the same lake in different years. On the contrary, for practical reasons we derive a fixed set of parameters for all winters and the whole (admittedly, much smaller) set of target lakes.
\par
Compared to MODIS, the literature on lake ice monitoring with VIIRS is limited. \cite{Suetterlin2017} estimated the LIP dates for winter 2016-17 in selected Swiss lakes using the LSWT derived from visible and near-infrared reflectances, and VIIRS thermal infrared band ($I_{5}$). Later, for winter 2016-17, \cite{Tom2020_lakeice_RS_journal} estimated the LIP dates of lakes Sihl, Sils, Silvaplana, and St.~Moritz from VIIRS and MODIS data. To our knowledge, no multi-winter VIIRS-based LIP trend analysis has been reported yet.
%\par
To summarise, most related works reviewed so far have found trends towards later freeze-up, earlier break-up and declining freeze duration. The prevalent methods are physics-inspired models based on empirical indices and thresholds. To our knowledge, none of the earlier trend studies applied Machine Learning (ML) methods to identify lake ice. \cite{lakeice_MODIS_Tibet2013} used k-means clustering to group the target lakes but not to detect lake ice.
%\par
%
\vspace{-1.25em}
\subsection{Lake ice observation with machine learning}
\vspace{-0.5em}
The last decades have seen the rise of ML in remote sensing and the Earth sciences. That is, large-scale statistical data analysis is used to capture the complex input-output relationships in a data-driven manner.
In \cite{Tom2020_lakeice_RS_journal}, we investigated pixel-wise classification of the spatio-temporal extent of lake ice from MODIS and VIIRS imagery with SVM. Each pixel was classified as either \textit{frozen} or \textit{non-frozen} in a supervised manner. We presented extensive experiments on data from two full winters and confirmed the efficacy of SVM for lake ice monitoring with MODIS and VIIRS. In this study, we apply SVM to quantify the 20-year lake ice trends. 
\cite{muyan_lakeice_2018} and \cite{rajanie_tom_2020} explored the potential of convolutional neural networks for lake ice detection in terrestrial webcam images (RGB). They performed a supervised classification of the lake pixels using the Tiramisu~\citep{Jegou2016_CVPRW}, respectively Deeplab v3+~\citep{deeplabv3plus2018} networks, into the four classes: \textit{water, ice, snow} and \textit{clutter}. 
Recently, \cite{Hoekstra2020} proposed an automated approach for ice vs.\ water classification in RADARSAT-2 data, combining unsupervised iterative region growing using semantics and supervised random forest labelling. A deep learning approach to lake ice detection in Sentinel-1 SAR imagery has been described in~\cite{tom_aguilar_2020}, and achieved promising results, including transferability across lakes and winters. Very recently, \cite{Wu_Duguay_2020_RSE} compared the capabilities of four different ML methodologies: multinomial logistic regression, SVM, random forest, and gradient boosting trees for lake ice observation using the MODIS data. They modelled lake ice monitoring as a 3-class (\textit{ice, water, cloud}) supervised classification problem. The four classifiers were tested on 17 large lakes from North America and Europe with areas $>1040~km^{2}$, and achieved \textgreater94\% accuracy. Random forest and gradient boosting trees showed better generalisation performance on this dataset of large lakes. 
\vspace{-1.5em}
\subsection{Definitions used}
\vspace{-0.5em}
The pixels that lie completely inside a lake are termed as \textit{clean pixels}. \textit{Non-transition dates} are the days when a lake is either completely frozen or completely non-frozen, the remaining days in a winter season are termed \textit{transition dates}. For each winter, we process all the dates from the beginning of September until the end of May. Definitions of the key LIP events are shown in Table \ref{table:phenology}.
\begin{table}[th]
\small
	    \centering
	    \begin{tabular}{cl} 
		\toprule
		\textbf{Event}  & \textbf{Definition}\\ 
		\midrule
	    FUS & 30\% or more of the non-cloudy lake portion\\
	    & is frozen and the just previous non-cloudy\\
	    & day should be $<30\%$ frozen
	    \\ 
		FUE & 70\% or more of the non-cloudy lake portion\\ & is frozen and the just previous non-cloudy\\
		& day should be $<70\%$ frozen\\
		BUS & 30\% or more of the non-cloudy lake portion\\
		& is non-frozen and the just previous non-cloudy\\
		& day should be $<30\%$ non-frozen
		\\ 
		BUE & 70\% or more of the non-cloudy lake portion\\
		& is non-frozen and the just previous non-cloudy\\
		& day should be $<70\%$ non-frozen
		\\ 
		\midrule
		ICD & BUE - FUS\\ 
		CFD & BUS - FUE\\ 
		\bottomrule
	    \end{tabular}
	    \caption{Key LIP events: Freeze-Up Start (FUS), Freeze-Up End (FUE), Break-Up Start (BUS), Break-Up End (BUE), Ice Coverage Duration (ICD) and Complete Freeze Duration (CFD).}
	    \label{table:phenology}
	    \normalsize
	\end{table}%
\vspace{-1.5em}
\section{Study area and data}\label{sec:data}
\vspace{-0.25em}
\subsection{Study area}\label{sec:method:study_area}
\begin{figure}[th]
\centering
  \includegraphics[width=0.99\linewidth]{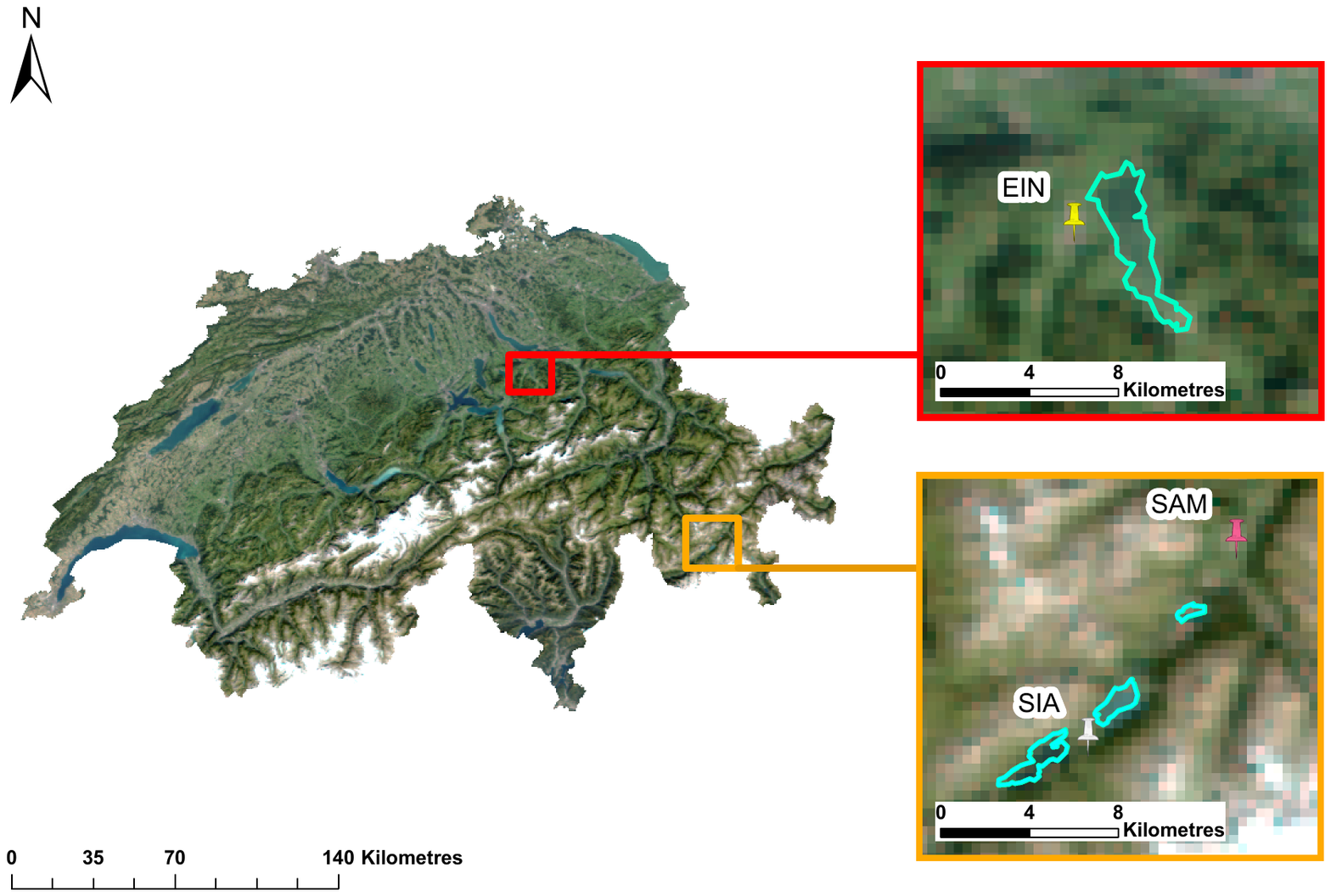}
  \caption{MODIS orthophoto map (RGB composite, red: $B_{1}$, green: $B_{4}$, blue: $B_{3}$) of Switzerland (left) captured on 7 September 2016. Red and amber rectangles show the regions Einsiedeln (around lake Sihl) and Oberengadin (with lakes Sils, Silvaplana and St. Moritz, from left to right) respectively. Inside each zoomed rectangle on the right, the respective lake outlines are shown in light green and the nearest meteorological stations (EIN:Einsiedeln, SIA: Segl Maria, SAM: Samedan) are marked using pins.} 
\label{fig:title_figure_lakes}
\end{figure}

\vspace{-0.25em}
We process four small- to medium-sized Swiss Alpine lakes: Sihl ($11.3~km^{2}$), Sils ($4.1~km^{2}$), Silvaplana ($2.7~km^{2}$) and St. Moritz ($0.8~km^{2}$), see Fig. Fig.~\ref{fig:title_figure_lakes} and Table~\ref{table:target_lakes}.
For the three small lakes in the region Oberengadin (Sils, Silvaplana, St.~Moritz), located at an altitude $>1750~m$, there are long in-situ observation series (important for climate studies), and they are also included in the NSIDC lake ice database (\url{https://nsidc.org}), although not updated recently. The fourth lake (Sihl) from the region Einsiedeln is relatively larger, lies at a lower altitude on the North part of the Alps, and has different environmental conditions.
\begin{table}[th]
	    \centering
	    \begin{tabular}{ccccc} 
		\toprule
		\textbf{} & \textbf{Sihl}  & \textbf{Sils}  & \textbf{Silva-} & \textbf{St.} \\  &   &   & \textbf{-plana} & \textbf{Moritz} \\ \midrule
		Latitude ($^{\circ}N$) & 47.14 & 46.42 & 46.45 & 46.49 \\ 
		Longitude ($^{\circ}E$) & 8.78 & 9.74 & 9.79 & 9.85 \\ Altitude ($m$) & 889 & 1797 & 1791 & 1768 \\
	    Max. depth ($m$) & 23 & 71 & 77 & 42\\
	    Avg. depth ($m$) & 17 & 35 & 48 & 26\\
	    Area ($km^\text{2}$) & 11.3 & 4.1  & 2.7 & 0.78 \\
		Volume ($Mm^\text{3}$) & 96 & 137 & 140 & 20\\
		\midrule
	    Meteo station & EIN
	    & SIA & SIA & SAM\\
		Latitude ($^{\circ}N$) & 47.13 & 46.43 & 46.43 & 46.53 \\
		Longitude ($^{\circ}E$) & 8.75 & 9.77 & 9.77 & 9.88 \\
		Altitude ($m$) & 910 & 1804 & 1804 & 1708 \\
	    \bottomrule
	    \end{tabular}
	    \caption{Details of the lakes (primary source: Wikipedia). The last four rows display information about the nearest meteorological stations.}
	    \label{table:target_lakes}
	\end{table}%

\par
\begin{figure}[t]
\centering
  \includegraphics[width=1.0\linewidth]{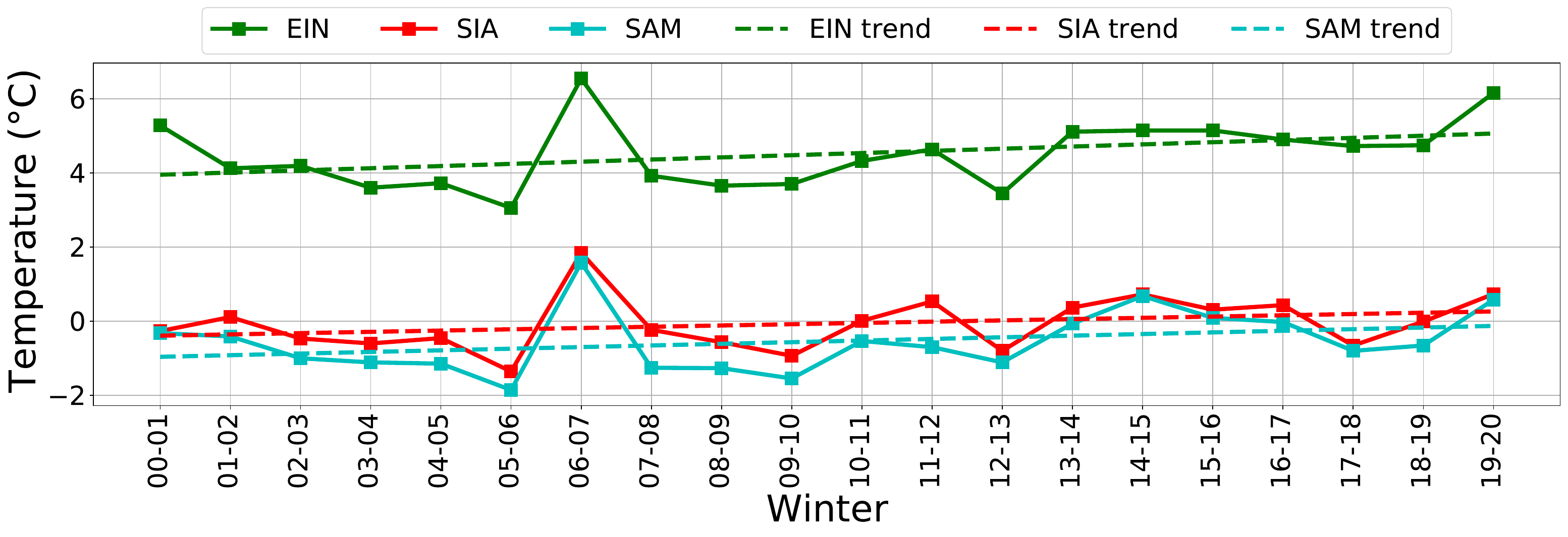}\\
  \includegraphics[width=1.0\linewidth]{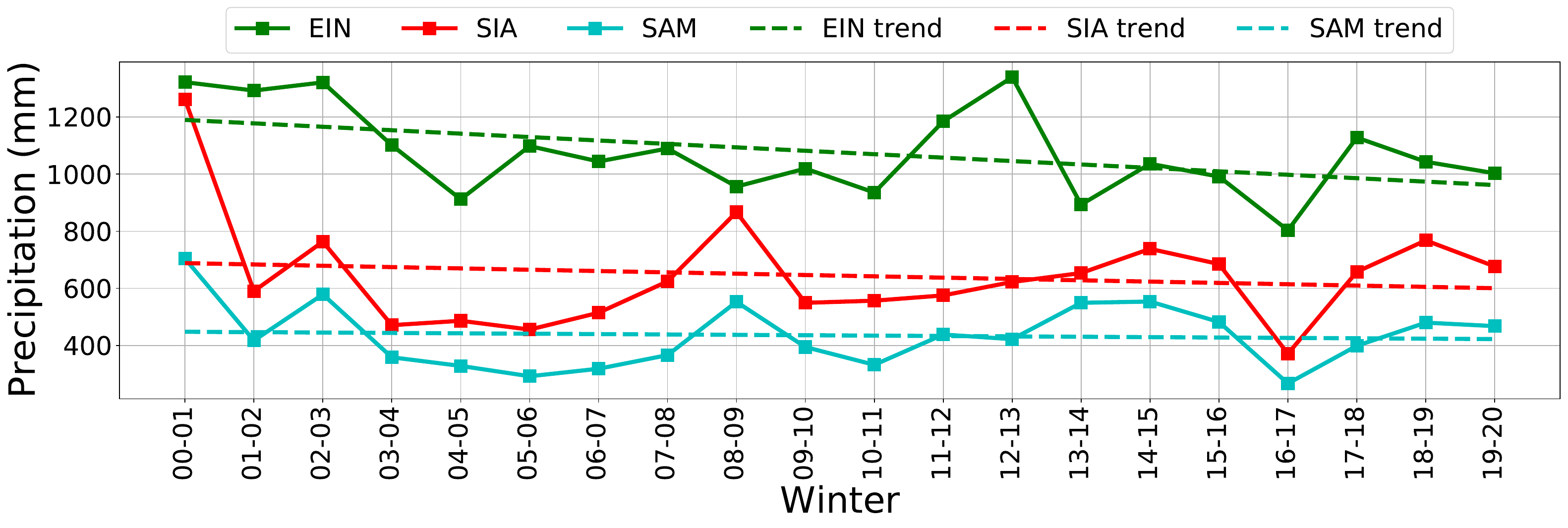}
\caption{Mean winter air temperature (row 1) and total winter precipitation (row 2) are plotted (solid curve, y-axis) against the winters shown in chronological order (x-axis). Twenty winter data from the nearest meteorological stations: EIN (Sihl), SIA (Sils and Silvaplana) and SAM (St.~Moritz) are used. The corresponding trends (linear fit, dotted curve) are also shown with the same colour. Data courtesy of MeteoSwiss. Winter 00-01 represents the dates from September 2000 till May 2001 (and similarly other winters).}
\label{fig:20winter_temperature}
\end{figure}

For these four lakes, there is no reference freeze/thaw data available from the past two decades. Hence, we study the weather patterns in the regions near the lakes. For each lake, the temperature and precipitation data recorded at the nearest meteorological stations are shown in Fig.~\ref{fig:20winter_temperature}. 
It can be seen that during the past 20 winters, at all the three meteorological stations, the mean temperature follows an increasing trend. On the other hand, precipitation has a decreasing pattern. While Meteoswiss has reported a significant trend of temperature increase in the Swiss Alps since 1864, they have so far not confirmed a notable precipitation trend (\url{https://www.meteoswiss.admin.ch/home/climate/climate-change-in-switzerland/temperature-and-precipitation-trends.html}). Over the shorter period of the past 20 winters, precipitation has been slightly declining. Warmer winters at higher altitudes in Switzerland could be linked to a decrease in precipitation, see \cite{Rebetez_1996}. The pattern of precipitation over the 20 years differs somewhat between the station EIN and the two other (similar) stations, e.g., see the winters 2008-09, 2012-13.
\vspace{-1.5em}
\subsection{Data}
\label{sec:Data}
\vspace{-0.25em}
In our analysis, we use the data from Terra MODIS~(\url{https://terra.nasa.gov/about/terra-instruments/modis}) and Suomi NPP VIIRS~(\url{https://ncc.nesdis.noaa.gov/VIIRS/}) satellites downloaded from the LAADS (\url{https://ladsweb.modaps.eosdis.nasa.gov}) and NOAA~(\url{https://www.avl.class.noaa.gov/}) databases, respectively. For MODIS processing, we downloaded the MOD02 (geolocated and calibrated radiance, level 1b, Top Of Atmosphere), MOD03 (geolocation) and MOD35\_L2 (cloud mask) products and pre-processed using \textit{MRTSWATH}~(\url{https://lpdaac.usgs.gov/tools/modis_reprojection_tool_swath/}, re-projection and re-sampling) and \textit{LDOPE}~\citep[cloud mask]{LDOPE} software. For VIIRS, we downloaded the Scientific Data Record data for the imagery bands, IICMO and VICMO products for the cloud masks, and GITCO (for image bands) and GMTCO (for cloud masks) for terrain corrected geolocation. VIIRS pre-processing is done using the following software packages: \textit{SatPy}~(\url{https://satpy.readthedocs.io/}) for assembling the data granules, mapping and re-sampling, \textit{H5py}~(\url{https://www.h5py.org}) for cloud mask extraction, \textit{PyResample}~(\url{https://resample.readthedocs.io}) and \textit{GDAL}~(\url{https://gdal.org}) for re-sampling of cloud masks. 
\par
\begin{figure}[t]
\centering
  \subfloat[MODIS pixels]{\includegraphics[width=0.65\linewidth]{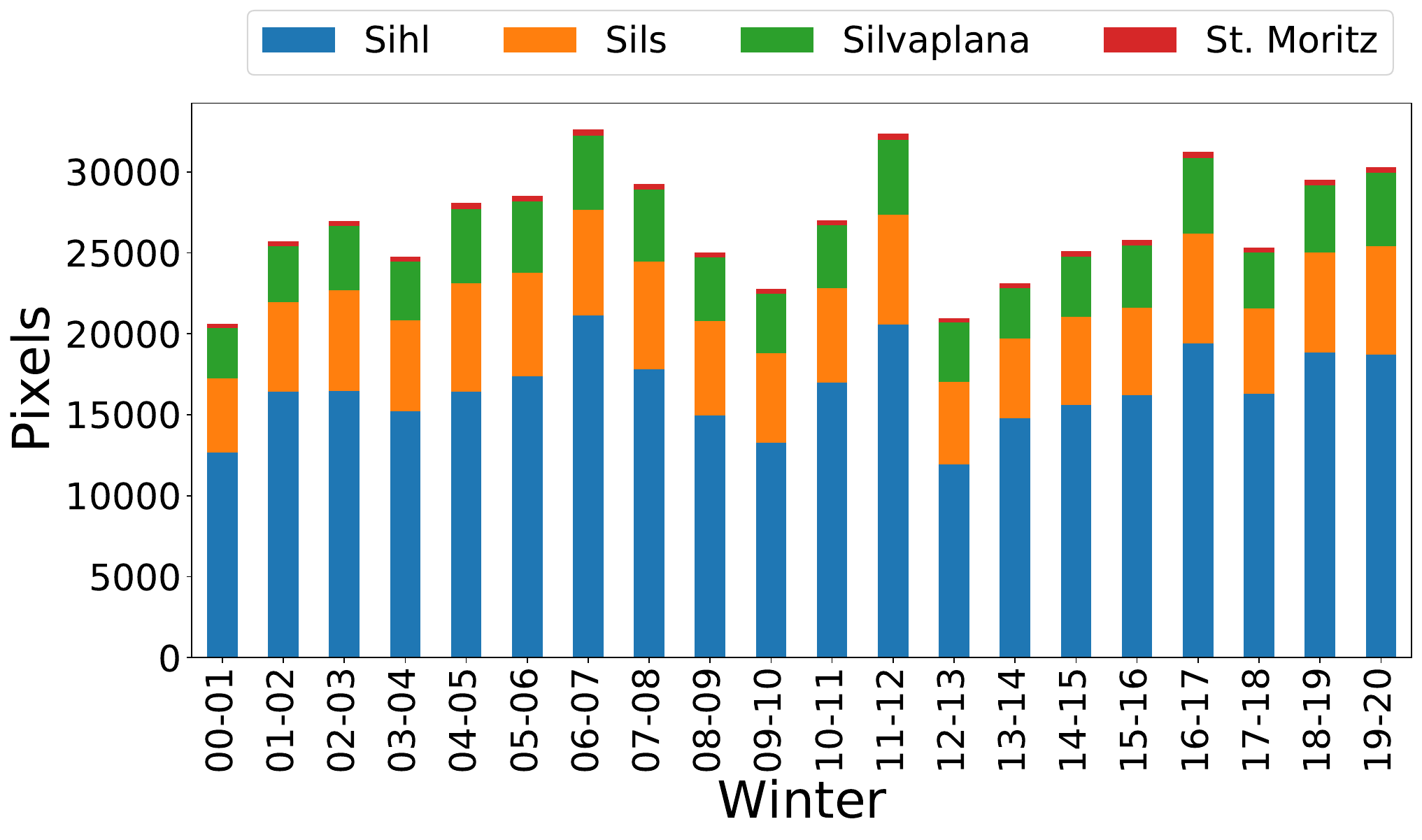}} 
  \subfloat[VIIRS pixels]{\includegraphics[width=0.350\linewidth]{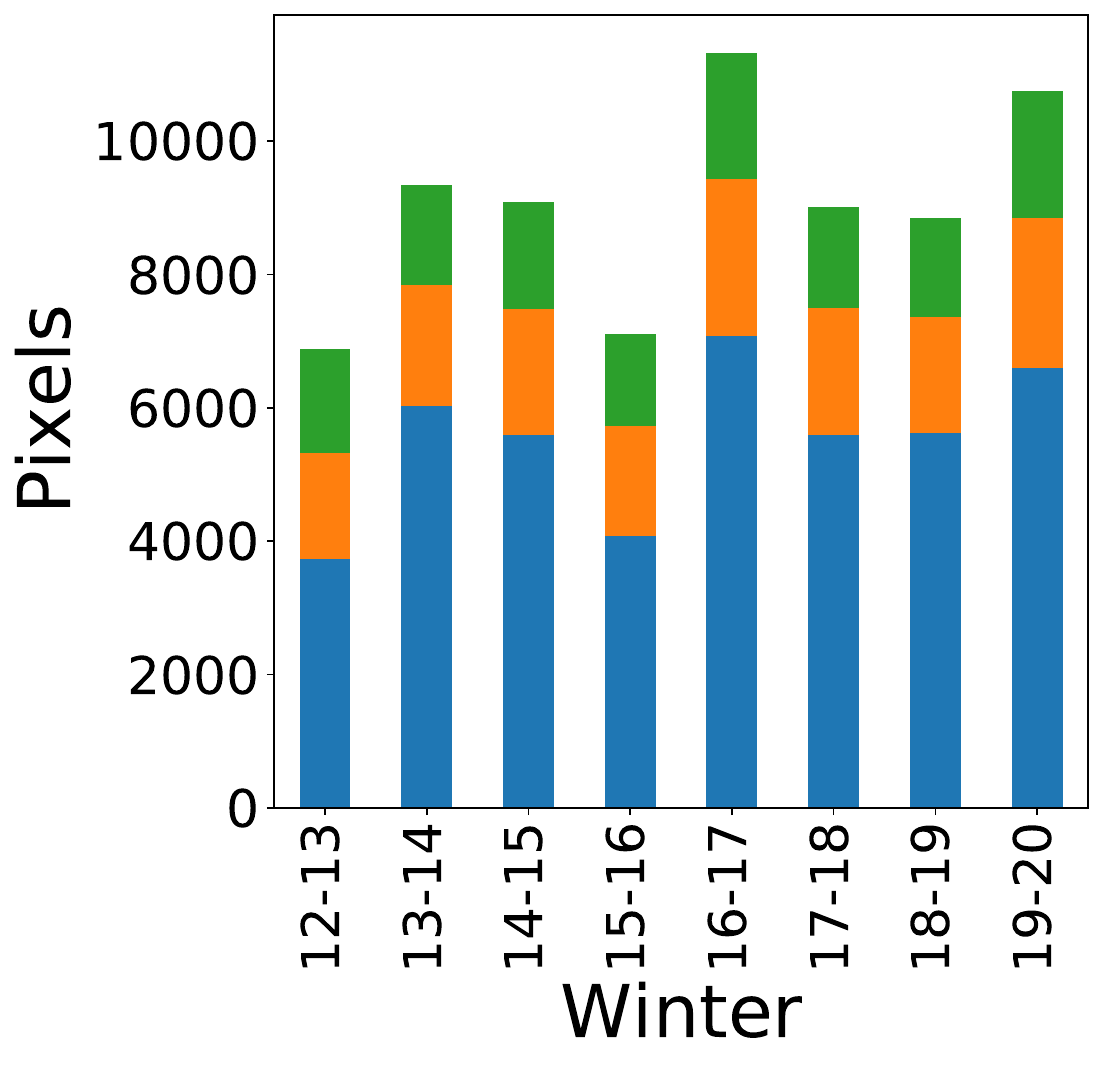}}\\
  \subfloat[MODIS clouds (20 winters)]{\includegraphics[width=1.0\linewidth]{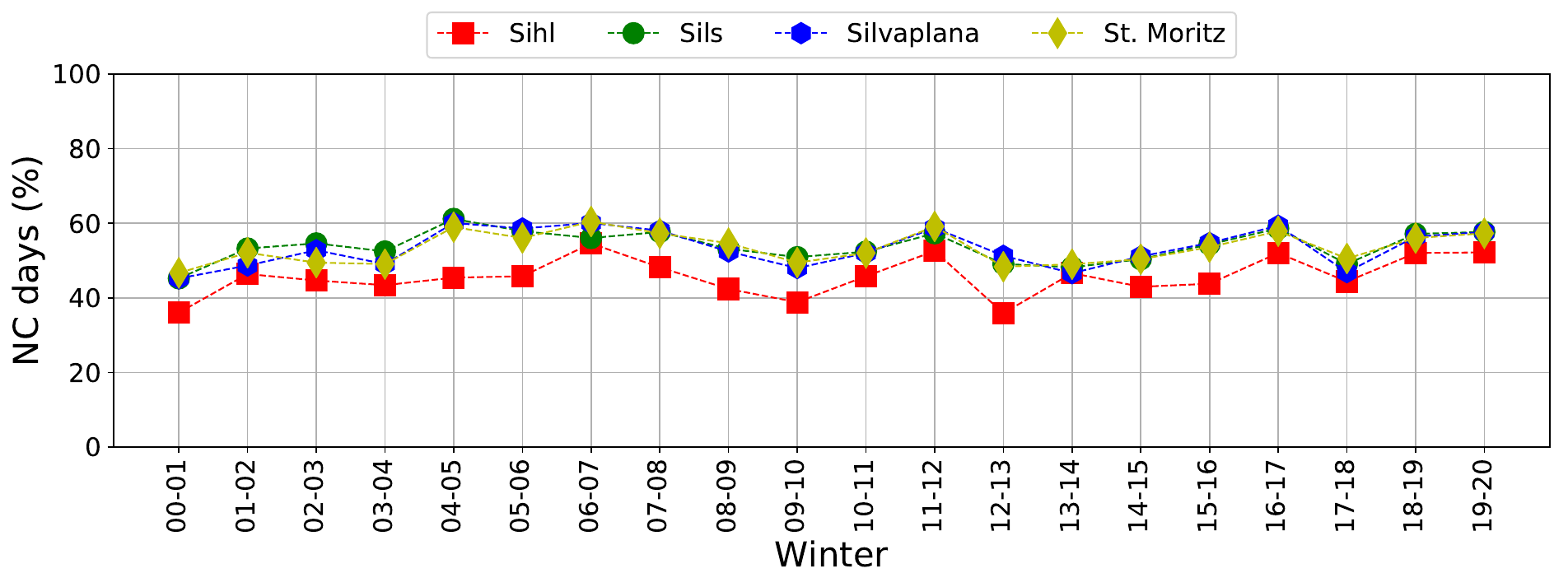}}
  \caption{Row 1 displays the \textit{clean, cloud-free} pixels (\textit{transition} and \textit{non-transition} dates) from the four target lakes. Data from both MODIS (20 winters, 4 lakes) and VIIRS (8 winters, 3 lakes) is displayed. Row 2 shows the percentage of at least 30\% Non-Cloudy (NC) days during each winter.} 
\label{fig:20winter_pixelstats}
\end{figure}

Fig.~\ref{fig:20winter_pixelstats} displays more details of the data that we use as a stacked bar chart (one colour per lake). For all target lakes, the total number of pixels in each winter is shown on the $y$-axis, against the winters in chronological order on the $x$-axis. In Fig.~\ref{fig:20winter_pixelstats}, note that some winters are relatively less cloudy and hence the number of cloud-free pixels varies across winters, even for the same lake and sensor. 
Due to its small size there exist no clean pixel for St.~Moritz in VIIRS imagery bands~\citep{Tom2020_lakeice_RS_journal}, hence we exclude it from the VIIRS analysis. 
It can be inferred from Fig.~\ref{fig:20winter_pixelstats} that, contrary to lake Sihl located at a lower altitude with different surrounding topography, the cloud patterns of lakes Sils, Silvaplana and St.~Moritz are quite similar, due to geographical proximity (see also Fig.~\ref{fig:title_figure_lakes}). However, minor differences exist (in a few winters) between the two very nearby lakes Sils and Silvaplana due to cloud mask errors. Different acquisition times of MODIS and VIIRS within a day can also result in varying cloud masks. Fig.~\ref{fig:20winter_pixelstats}c shows that approximately 40 to 60\% of all observations per winter are covered by clouds, strongly reducing the effective temporal resolution of the time series.

\par
\textbf{Ground truth}. Our ground truth is based on the visual interpretation of freely available high-resolution images from webcams monitoring the target lakes. One label (\textit{fully frozen, fully non-frozen, partially frozen}) per day is assigned. Two different operators looked at each image, i.e., a second expert verified the judgement of the first operator to minimise interpretation errors. When deciphering a webcam image was difficult, additional images were used from other webcams viewing the same lake (if available), images from the same webcam but at other acquisition times on the same day, and images of the same webcam for the days before and after the given observation day. We also improved the webcam-based ground truth using sporadic information from media reports, and by visually interpreting Sentinel-2 images, whenever available and cloud-free. No webcam data is available from the winters before 2016-17. Moreover, the manual interpretation process is labour intensive. Thus, ground truth is available only for winters 2016-17 and 2017-18. 
Even though visual interpretation is the standard practice, a certain level of label noise inevitably remains in the ground truth, due to factors such as interpretation errors, image compression artefacts, large distance and flat viewing angle on the lake, etc. Furthermore, the webcams used are not optimally mounted for lake ice monitoring and hence do not always cover the full lake area (or even a major portion of it), even for the smallest lake St.~Moritz. 
Still, the ground truth serves the purpose, in the sense that it has significantly fewer wrongly labelled pixels than the automatic prediction results. For the winters 2016-17 and 2017-18, we see no possibility to obtain a more accurate, spatially explicit ground truth for our task.
\vspace{-1.25em}
\section{Methodology}
\begin{figure}[t]
\centering
  \includegraphics[width=0.8\linewidth]{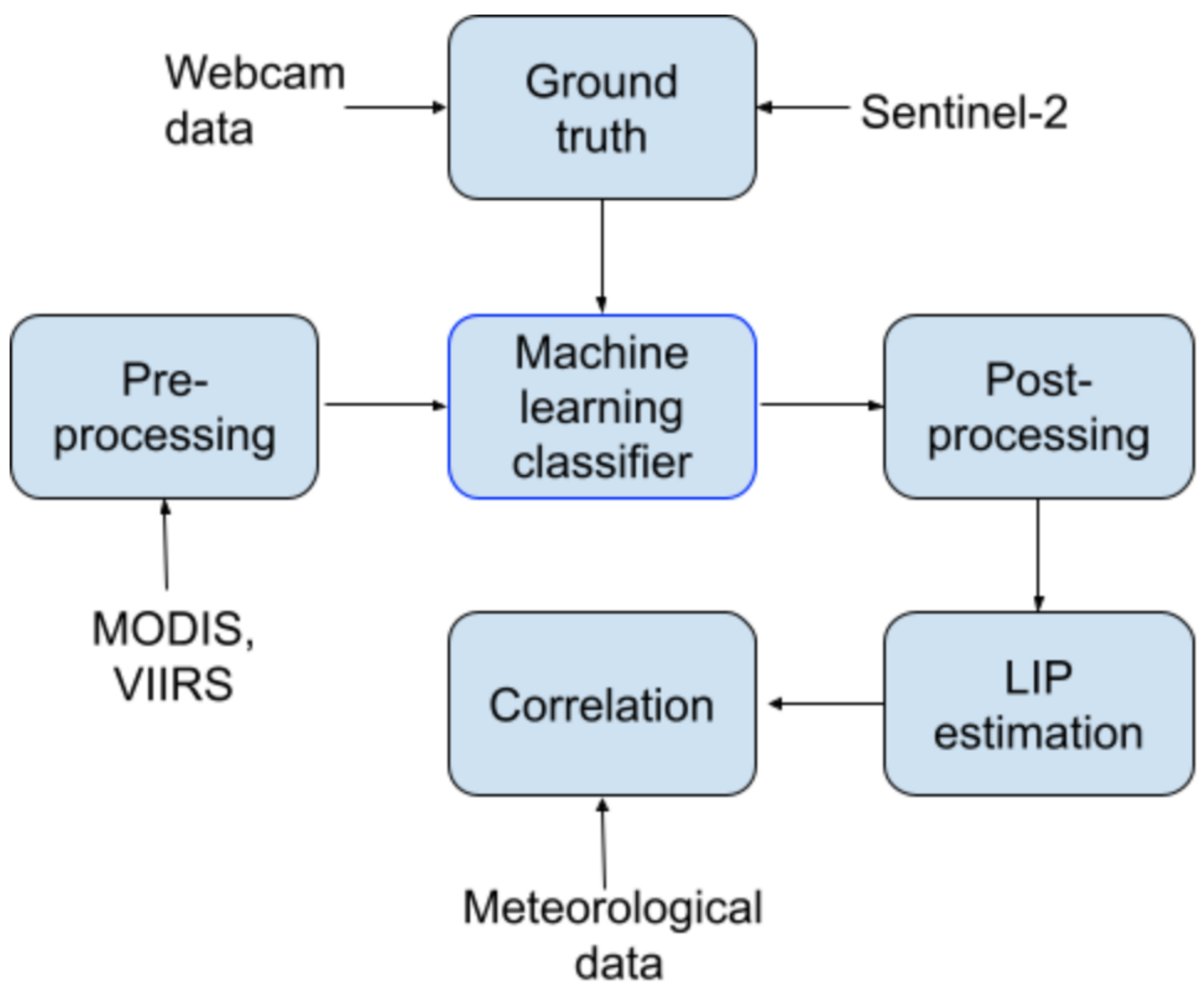}
  \caption{Flow diagram of the proposed methodology.}
\label{fig:flow_diagram}
\end{figure}

\vspace{-0.25em}
A flowchart of the method is shown in Fig.~\ref{fig:flow_diagram}.
We perform the pre-processing steps as in \cite{Tom2020_lakeice_RS_journal}. First, the absolute geolocation error for both sensors (0.75, respectively 0.85 pixels $x$- and $y$-shifts for MODIS; 0.0, respectively 0.3 pixels $x$- and $y$-shifts for VIIRS) are corrected. The generalised \citep{DouglasPeucker73} lake outlines are then back-projected onto the images to extract the clean pixels. Mixed pixels are discarded from the analysis. Binary cloud masks are derived from the respective cloud mask products to limit the analysis only to cloud-free pixels. We super-resolved all low resolution MODIS bands ($500~m$, $1000~m$) to $250~m$ using bilinear interpolation prior to the analysis. This step is not required for VIIRS as all used bands have the same GSD ($\approx375~m$).
\vspace{-1.5em}
\subsection{Machine learning for lake ice extraction}
\label{ML_methodology}
\vspace{-0.25em}
We model lake ice detection in optical satellite images as a per-pixel 2-class (\textit{frozen, non-frozen}) supervised classification problem and employ a linear SVM~\citep{SVM} classifier. As in \cite{Tom2020_lakeice_RS_journal}, for each pixel, the feature vector is formed by directly stacking the 12 (5) bands of MODIS (VIIRS). The bands that offer maximum separability for the task of lake ice monitoring were automatically chosen by the supervised XGBoost feature selection algorithm~\citep{xgboost}. We treat \textit{snow-on-ice} and \textit{snow-free-ice} as a single class: \textit{frozen}. Class \textit{non-frozen} denotes the open water pixels.  %
\par
While there recently has been a strong interest in deep learning for remote sensing tasks~\citep{CampsValls_wiley_2021}, deep neural networks are not suitable for our particular application, due to the scarcity of pixels with reliable ground truth. The lakes that we monitor are small and ground truth is available only for two winters (see Section~\ref{sec:Data}), which is too little to train data-hungry neural networks. Also, given the large GSD and limited need for spatial context, we do not expect deep models to greatly outperform shallower ones.
\vspace{-1.5em}
\subsection{LIP estimation}
\label{sec:LIP_estimation}
\begin{figure}[t]
\centering
  \subfloat[Results timeline before curve fitting]{\includegraphics[width=1.0\linewidth]{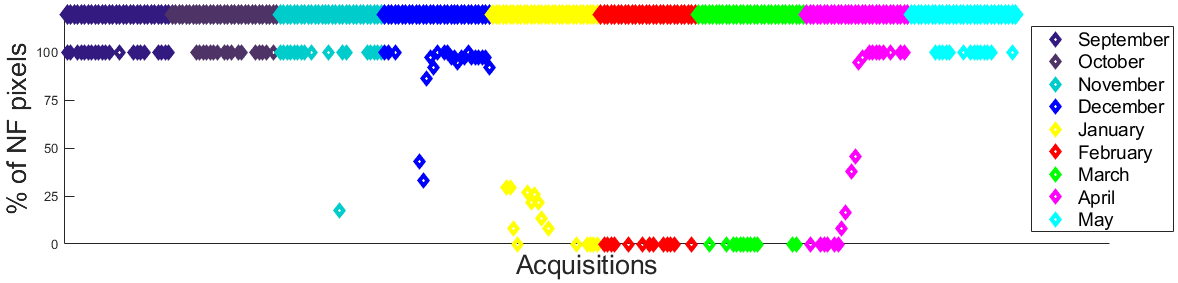}} \\
  \subfloat[After curve fitting]{\includegraphics[width=1.0\linewidth]{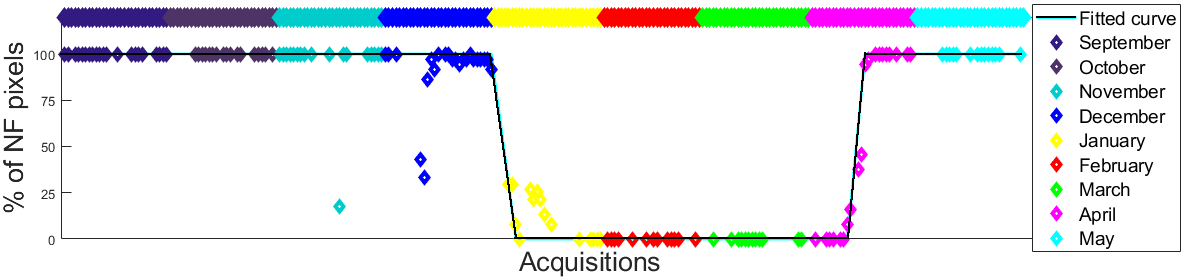}} \caption{Piece-wise linear ("U with wings") curve fitting example. NF indicates Non-Frozen. Results are displayed only for the cloud-free acquisitions.}
\label{fig:LIP_curve_method}
\end{figure}

\vspace{-0.25em}
Each winter, using the trained ML model, we process all available non-cloudy acquisitions and generate pixel-wise classification maps (one per acquisition). To recover the temporal evolution (per winter), the percentage of non-frozen pixels is computed from each classification map and is plotted on the $y$-axis against the acquisition time on the $x$-axis. Then, as in \cite{Tom2020_lakeice_RS_journal}, multi-temporal smoothing is performed using a Gaussian kernel with a standard deviation of $0.6$ days and window width of $3$ days. An example MODIS results (timeline) for lake Sils from winter 2006-07 is shown in Fig.~\ref{fig:LIP_curve_method}a. Results from different months are displayed in different colours, see the legend. 
\par
After multi-temporal smoothing, we find all the potential candidates for the following four critical dates: FUS, FUE, BUS and BUE, see Table~\ref{table:phenology} for the corresponding definitions. 
Within a winter, it is possible that $>\!$~1 candidates exist per critical date which all satisfy the respective definition. To weed out some spurious candidates, we enforce the constraint that the four dates must occur in the following chronological order: FUS$\rightarrow$FUE$\rightarrow$BUS$\rightarrow$BUE.
Then we exhaustively search for the optimal set of four dates among the remaining candidates. To that end, we fit a continuous, piece-wise linear \textit{"U with wings"} shape to the per-day values of percentage of non-frozen pixels, such that the fitting residuals $z$ are minimised (see example fit in Fig.~\ref{fig:LIP_curve_method}b, shown in black colour). In detail, the loss function for the fit is defined as:
\begin{equation}
L_{LIP} = \frac{1}{P}\cdot \sum_{i=1}^{N} %H_{\phi}(\frac{z-\mu}{\sigma})
H_{\phi}(z)
\end{equation}
where $N$ is the total number of \textcolor{black}{cloud-free} acquisitions. 
\vspace{-0.75em}
\begin{equation}
H_{\phi}(z) =\begin{cases} 
z^{2}  & \mathopen|z\mathclose|  \leq  \phi\\
2\phi \mathopen|z\mathclose| - \phi ^{2}  & \mathopen|z\mathclose|  > \phi 
\end{cases}
\end{equation}
is the Huber norm of the residual. For the shape parameter $\phi$, we use a constant value of 1.35 which offers a good trade-off between the robust $l_1$-norm for large residuals and the statistically efficient $l_2$-norm for small residuals~\citep{Owen06arobust}.
\par
Per lake, we assume that each critical date occurs only once per winter, which is always true in Oberengadin. Lake Sihl does not always fully freeze. As it lies outside of the target region and is included mostly to ensure transferability of the ice classifier, we do not extract the LIP events for Sihl. Moreover, we decide to exclude lake St.~Moritz since it is too small for the GSD of MODIS (only 4 clean pixels), making the fraction of frozen pixels overly susceptible to noise. We thus prefer to study only the two main lakes in Oberengadin, Sils and Silvaplana, in terms of long-term lake ice trends. These two lakes fully freeze every year and typically have a single freeze-up and break-up period. To further stabilise the LIP estimates we include a weak prior probability for each phenological date, in the form of a diffuse Gaussian distribution. 
\par
The prior probability ($P$) is given by:
\begin{equation}
P = P_{fus} \cdot P_{fue}\cdot P_{bus} \cdot P_{bue}
\end{equation}
where $P_{fus}$, $P_{fue}$, $P_{bus}$, and $P_{bue}$ are Gaussian normal distributions for the events FUS, FUE, BUS and BUE, respectively. The prior formalises the knowledge that freeze-up normally occurs around the end of December and takes around three days, and break-up occurs around the end of April over a similar period, for both target lakes. To not bias the estimation, but only to minimise the risk of implausible results, we choose very wide Gaussians ($\sigma=1$ month). Furthermore, we impose a constraint that the duration of freeze-up (FUE-FUS) and break-up (BUE-BUS) is not more than two weeks. 
\par
We use $30\%$ as the threshold to estimate the four dates. For example, a date is considered a FUS candidate if 30\% or more of the non-cloudy portion of the lake is frozen. Some studies based on MODIS~\citep{Qinghai2020,Reed_2009,Yao2016} have used 10\% as the threshold, while another approach~\citep{lakeice_MODIS_Tibet2013} even employed 5\%. All of them monitored larger lakes ($45$ to $4294~km^{2}$ area). We empirically found that for our rather tiny lakes the above thresholds are too strict and a threshold of 30\% is needed to ensure reliable decisions. To see why, consider that in the best case (Sils, cloud-free) a lake has 33 clean pixels, but that number can go down to as few as 7 (Silvaplana, 70\% cloud cover). Note also that on such small lakes a large portion of all pixels is very close to the lake's shoreline, where the geo-location error (in the worst case 0.5 pixel) may have a significant impact.%
\vspace{-1.0em}
\section{Results}
\label{sec:EXPERIMENTS}
\vspace{-0.25em}
In addition to the overall classification accuracy we report also a stricter measure, mean intersection-over-union (mIoU), which better depicts the performance when the class distribution is imbalanced. Overall classification accuracy is given by:
\begin{equation}
Accuracy = \frac{TP + TN}{TP + TN + FP + FN}
\end{equation}
\vspace{0.1em}
where \textit{TP, TN, FP}, and \textit{FN} represent true positive pixels, true negatives, false positives and false negatives, respectively. For each class, the Intersection-over-Union score (IoU or Jaccard Index) is given by:
\begin{equation}
IoU = \frac{TP}{TP + FP + FN}
\end{equation}
\vspace{0.1em}
mIoU is the average of the per-class IoUs. 
\vspace{-1.25em}
\subsection{Experiments on MODIS data from 20 winters}
\label{section:modis_20winters}
\vspace{-0.25em}
\textbf{Test-train split}.
We process MODIS data from all the 20 winters since 2000-01 (inclusive).
Details of the training set for each tested winter are shown in Table~\ref{table:q20winters_train_set}.
To avoid systematic biases in the estimated ice maps due to overfitting to a particular year, we proceed as follows: we train the linear SVM model on all non-transition dates of 2016-17 and use it to estimate lake ice coverage for all days in 2017-18 (including transition dates). We repeat that procedure in the opposite direction, i.e., we train on all non-transition days of 2017-18 and perform inference for all dates of 2016-17. Then, we merge all non-transition dates from both winters into a new, larger \emph{two-winter} training set, which we further augment with an \emph{auxiliary dataset}. The latter contains all acquisitions of lakes Sils and Silvaplana captured during the remaining 18 years in September (when the lakes are never frozen) and in February (when the lakes are always frozen). The purpose of the auxiliary dataset is to cover a wider range of weather and lighting conditions that might not have been encountered in the two winters with annotated ground truth, for better \textcolor{black}{transferability}. Data of lake Sihl is not included in the auxiliary set, as it does not freeze reliably,  St.~Moritz is ignored due to its negligible number of pixels. The two-winter and auxiliary datasets are merged and used to train a linear SVM model, which is then used to predict ice cover maps for the 18 remaining winters.
\begin{table}[th]
\small
	    \centering
	    \begin{tabular}{ll} 
		\toprule
		\textbf{Test set} & \textbf{Training set}\\ 
		\midrule
		2016-17 & 2017-18\\ 
		2017-18 & 2016-17\\ 
		\textcolor{black}{remaining winters} & \textcolor{black}{2016-17, 2017-18, aux set} \\ 
		\bottomrule
	    \end{tabular}
	    \caption{\textcolor{black}{Test-train split for the MODIS data from 20 winters. Aux set refers to auxiliary set.}}
	    \label{table:q20winters_train_set}
	    \normalsize
	\end{table}
\par
\begin{figure}[t]
\centering
  %\subfloat[\textcolor{black}{8.11.00}]{%
  %    \includegraphics[width=0.11\textwidth]{T20001108_1135_Sihlsee_clas_mod_final.png}}\hspace{0.001em}
  %\subfloat[\textcolor{black}{1.2.02}]{%
   %   \includegraphics[width=0.11\textwidth]{T20020201_1110_Sihlsee_clas_mod_final.png}}\hspace{0.001em}
  %\subfloat[\textcolor{black}{29.4.03}]{%
  %    \includegraphics[width=0.11\textwidth]{T20030429_1045_Sihlsee_clas_mod_final.png}}\hspace{0.001em}
  %\subfloat[\textcolor{black}{29.12.03}]{%
  %    \includegraphics[width=0.11\textwidth]{T20031229_1020_Sihlsee_clas_mod_final26.png}}\\
  %\subfloat[\textcolor{black}{15.10.04}]{%
  %    \includegraphics[width=0.11\textwidth]{T20041015_1050_Sihlsee_clas_mod_final.png}}\hspace{0.001em}
  \subfloat[\textcolor{black}{1.1.2006}]{%
  \includegraphics[width=0.22\textwidth]{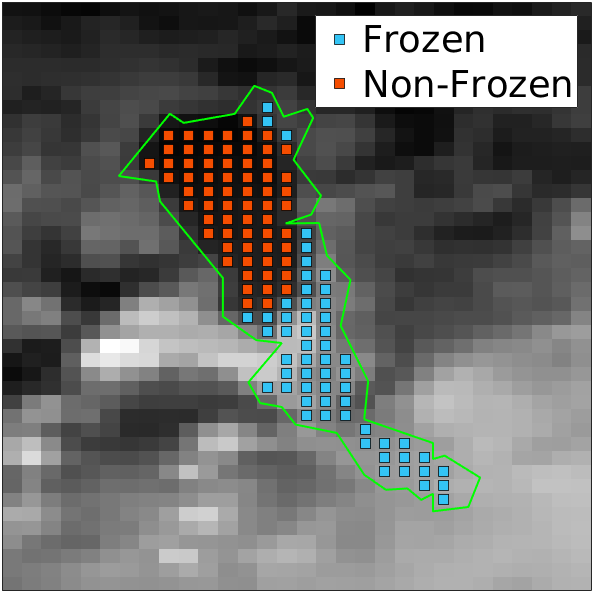}}\hspace{0.001em}
  %\subfloat[\textcolor{black}{30.3.07}]{%
  %    \includegraphics[width=0.11\textwidth]{T20070330_1050_Sihlsee_clas_mod_final.png}}\hspace{0.001em}
  %\subfloat[\textcolor{black}{8.2.08}]{%
  %    \includegraphics[width=0.11\textwidth]{T20080208_1030_Sihlsee_clas_mod_final.png}}\\
  %\subfloat[\textcolor{black}{9.1.09}]{%
  %    \includegraphics[width=0.11\textwidth]{T20090109_1030_Sihlsee_clas_mod_final26.png}}\hspace{0.01em}
  \subfloat[\textcolor{black}{23.1.2010}]{%
  \includegraphics[width=0.22\textwidth]{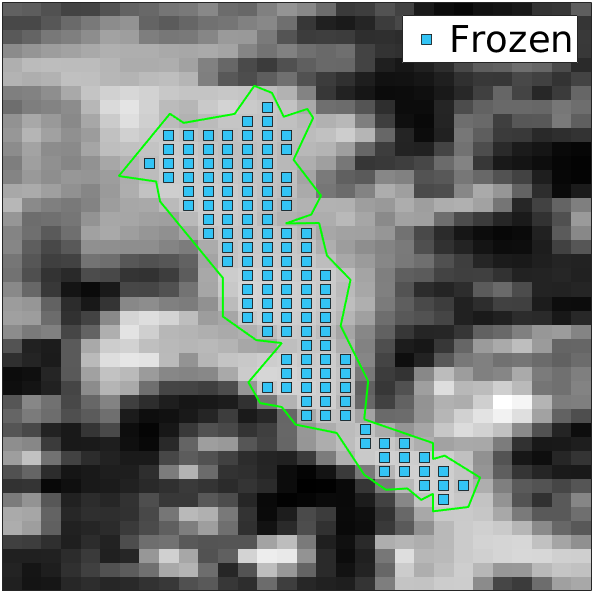}}\hspace{0.01em}\\
  %\subfloat[\textcolor{black}{24.3.11}]{%
  %    \includegraphics[width=0.11\textwidth]{T20110324_1005_Sihlsee_clas_mod_final.png}}\hspace{0.01em}
  %\subfloat[\textcolor{black}{2.3.12}]{%
  %    \includegraphics[width=0.11\textwidth]{T20120302_1055_Sihlsee_clas_mod_final.png}}\\
  %\subfloat[\textcolor{black}{24.4.13}]{%
  %    \includegraphics[width=0.11\textwidth]{T20130424_1045_Sihlsee_clas_mod_final.png}}\hspace{0.01em}
  \subfloat[\textcolor{black}{10.3.2014}]{%
  \includegraphics[width=0.22\textwidth]{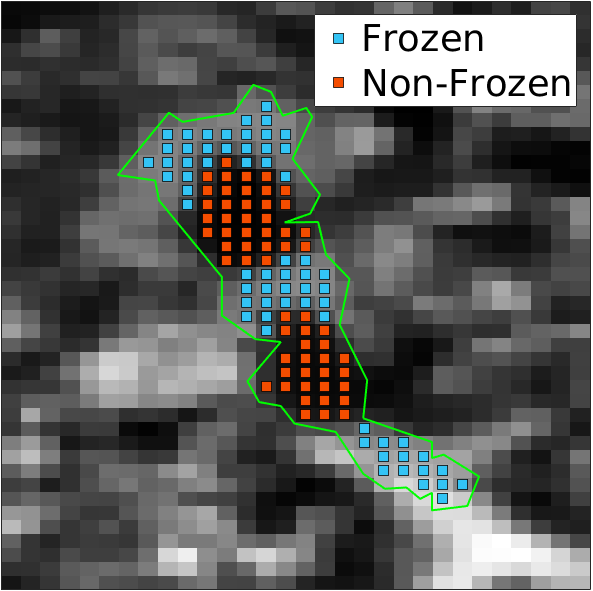}}\hspace{0.01em}
  %\subfloat[\textcolor{black}{28.1.15}]{%
  %    \includegraphics[width=0.11\textwidth]{T20150128_1020_Sihlsee_clas_mod_final.png}}\hspace{0.01em}
  %\subfloat[\textcolor{black}{22.1.16}]{%
  %    \includegraphics[width=0.11\textwidth]{T20160122_1025_Sihlsee_clas_mod_final.png}}\\
  %\subfloat[\textcolor{black}{15.5.17}]{%
  %    \includegraphics[width=0.11\textwidth]{T20170515_0945_Sihlsee_clas_mod_final.png}}\hspace{0.01em}
  %\subfloat[\textcolor{black}{21.4.18}]{%
  %    \includegraphics[width=0.11\textwidth]{T20180421_1000_Sihlsee_clas_mod_final.png}}\hspace{0.01em}
  %\subfloat[\textcolor{black}{5.2.19}]{%
  %    \includegraphics[width=0.11\textwidth]{T20190205_0950_Sihlsee_clas_mod_final.png}}\hspace{0.01em}
  \subfloat[\textcolor{black}{20.9.2019}]{%
  \includegraphics[width=0.22\textwidth]{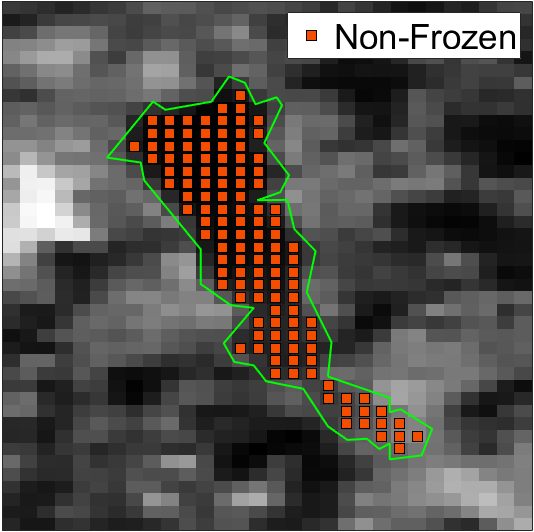}}\\
  \caption{\textcolor{black}{MODIS classification results (lake Sihl, overlaid on band $B_{1}$) on selected dates from the past 20 winters. blue and red squares are overlaid on the pixels detected as frozen and non-frozen respectively.}}
  \label{fig:MODIS_qualitative} 
\end{figure}

\textbf{Qualitative results}.
Exemplary qualitative results of lake Sihl are shown in Fig. \ref{fig:MODIS_qualitative}. The respective dates are displayed below each sub-figure. The lake outline overlaid on the MODIS band $B_{1}$ is shown in green. Pixels detected as frozen and non-frozen are shown as blue and red squares, respectively. The results include fully-frozen, fully non-frozen and partially frozen days.
\par
\begin{figure*}[t]
\centering
  \subfloat[]{\includegraphics[width=0.23\linewidth]{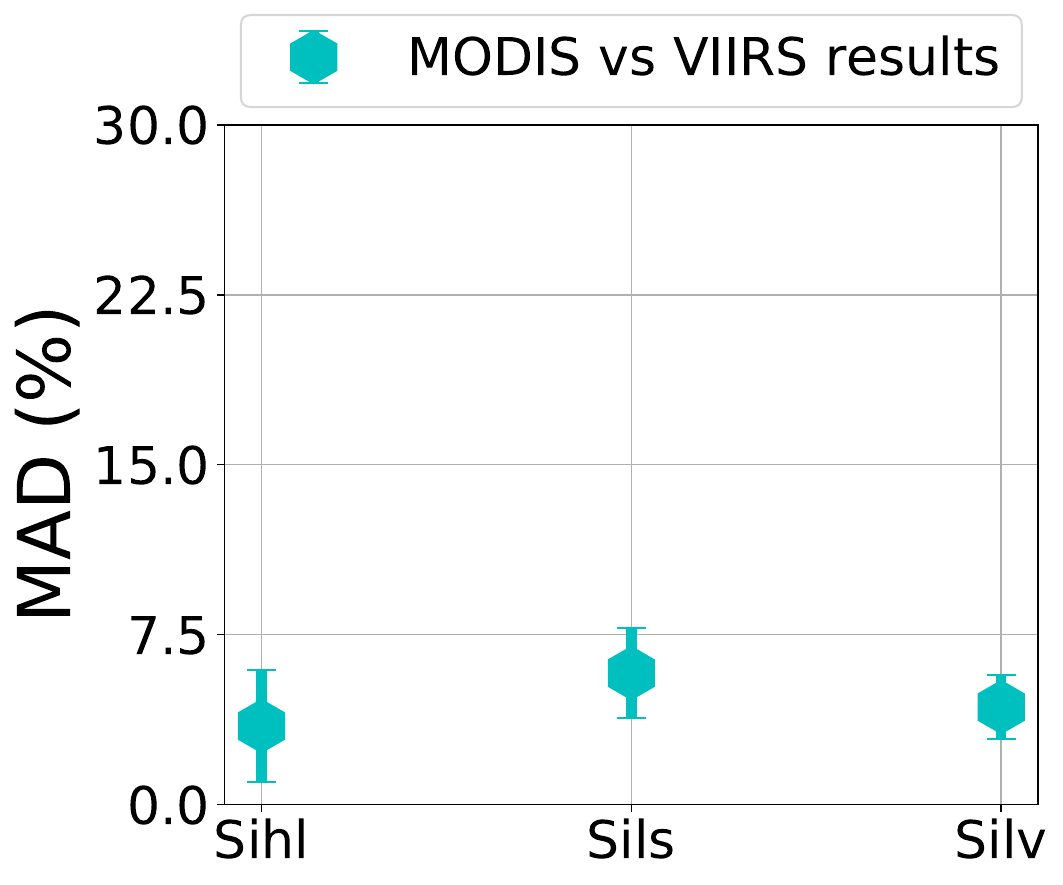}}
  \subfloat[]{\includegraphics[width=0.65\linewidth]{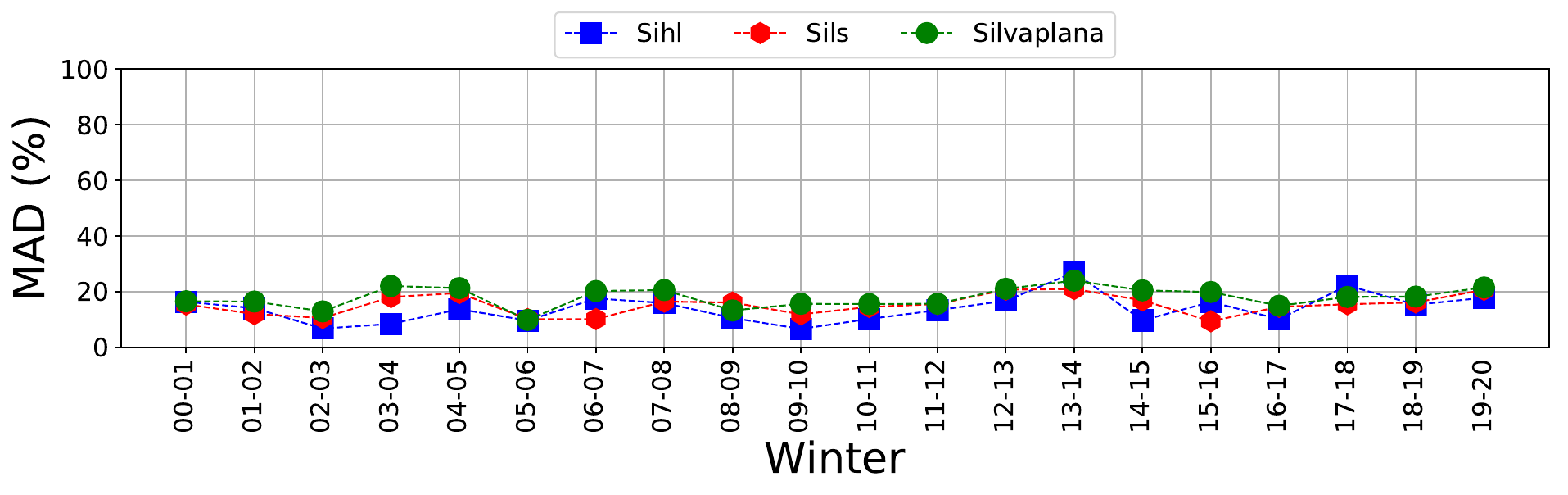}}\\
  \subfloat[]{\includegraphics[width=0.4\linewidth]{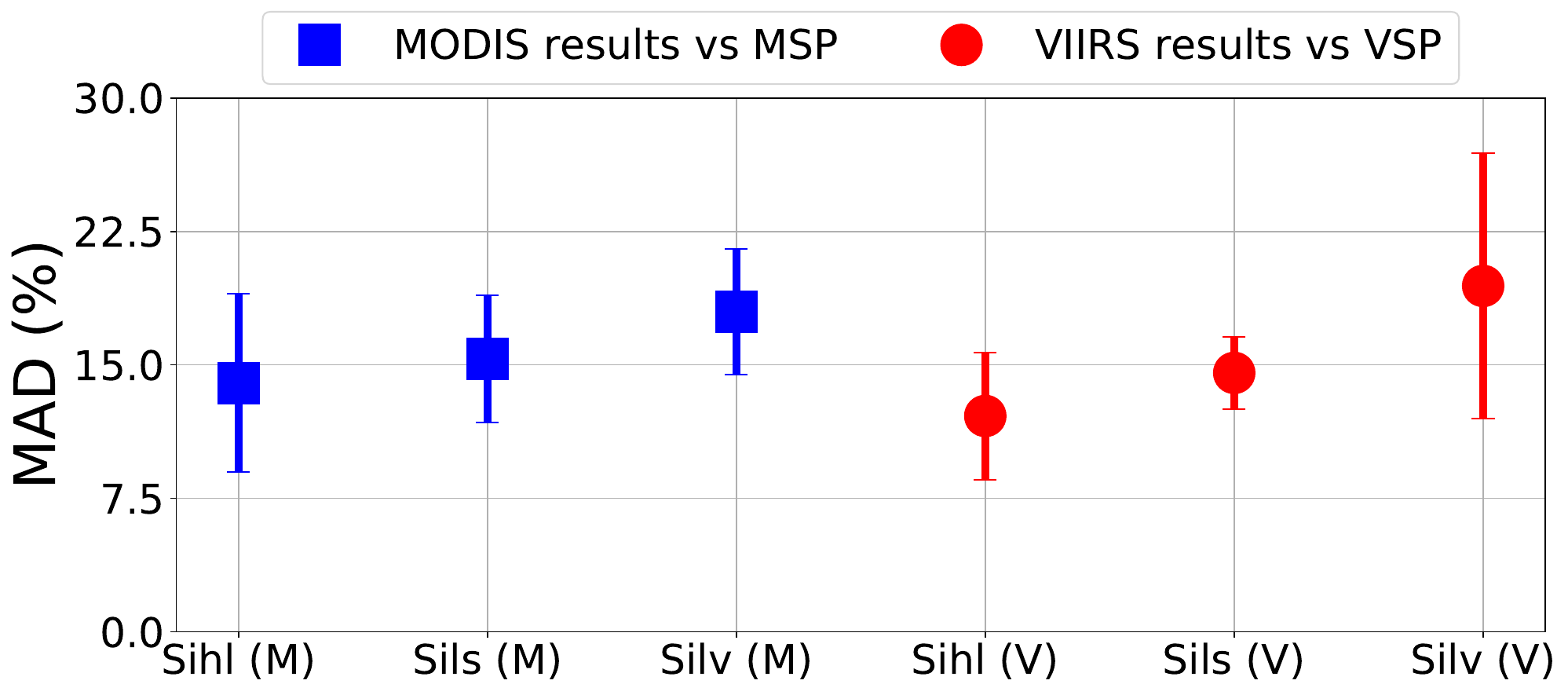}}
  \subfloat[]{\includegraphics[width=0.4\linewidth]{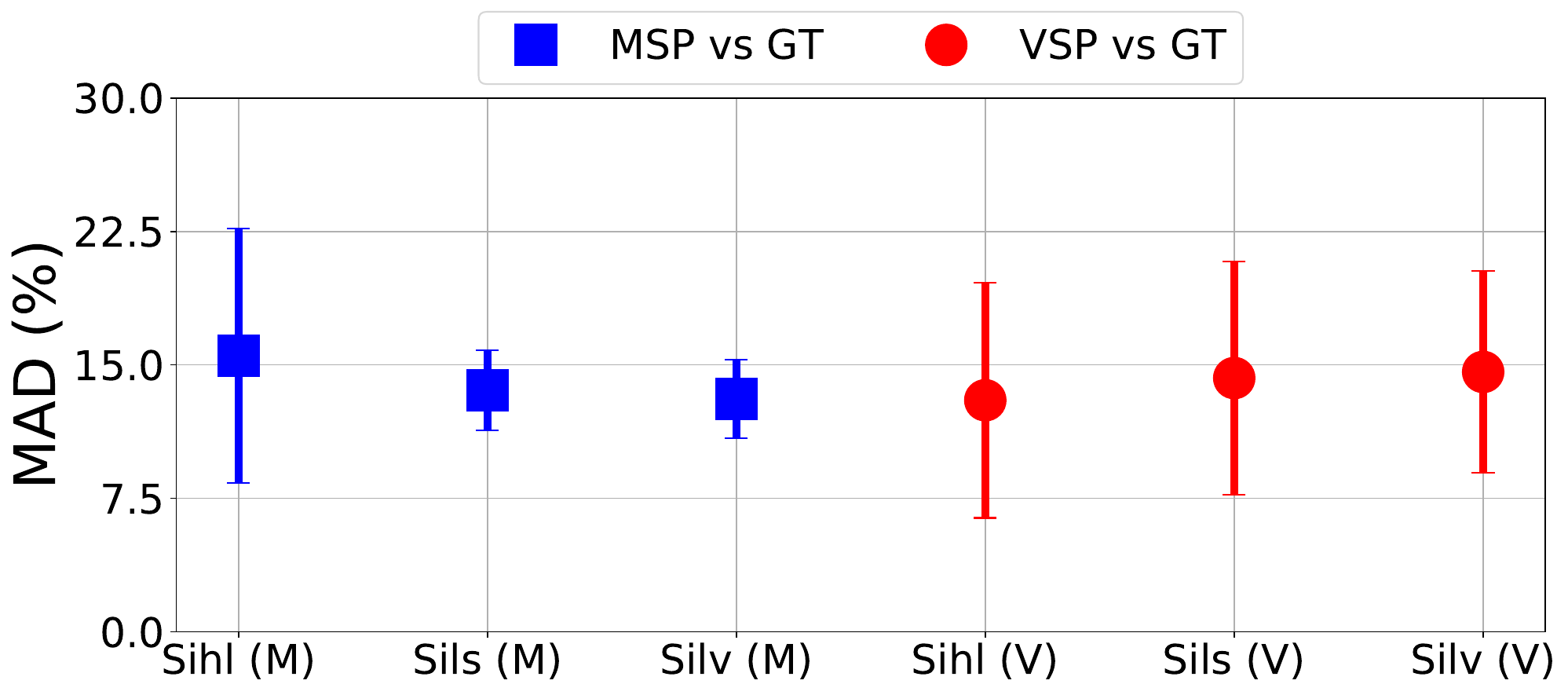}}
  \caption{\textcolor{black}{(a) The comparison of our \textcolor{black}{Machine Learning (ML) products} (MODIS and VIIRS) for the 8 common winters. MAD stands for Mean Absolute Difference. Silv represents lake Silvaplana. (b) Our \textcolor{black}{(MODIS) ML product} (per-winter MAD) vs.\ MODIS Snow Product (MSP) for 20 winters. (c) Comparison of our MODIS (M, 20 winters) and VIIRS (V, 8 winters) \textcolor{black}{ML products} with the respective snow products (MSP and VIIRS Snow Product, VSP). (d) Deviations between the two snow products and our webcam-based ground truth (GT).}}
\label{fig:comparison_MSP}
\end{figure*}

\vspace{-1.0em}
\textbf{Additional check using VIIRS}.
Direct quantitative analysis is not possible, since no ground truth is available for 18 out of the 20 winters. To validate our \textcolor{black}{ML product (MODIS)}, we additionally process the VIIRS data from 8 winters (since winter 2012-13, inclusive) and compare the results. Since a pixel-to-pixel comparison is not straightforward due to different GSDs, we fit the timelines per winter for each lake as described before (Fig.~\ref{fig:LIP_curve_method}a) and compute absolute differences (AD) between the daily estimates for the percentage of frozen pixels. The AD is computed only on dates when both MODIS and VIIRS acquisitions are present, and when the lake is at least 30\% cloud-free. The ADs are then aggregated to obtain a Mean Absolute Difference (MAD) per winter. Fig.~\ref{fig:comparison_MSP}a shows, for each lake, the mean and standard deviation of the MAD across the 8 common winters. The low mean values (3.5, 5.8 and 4.3 percent respectively for Sihl, Sils and Silvaplana) show that our MODIS and VIIRS \textcolor{black}{ML products} are in good agreement, especially considering that a part of the MAD is due to the difference in GSD between MODIS ($250~m$) and VIIRS ($\approx375~m$). Note also that the acquisition times during the day (and hence the cloud masks) can differ; and that, although the absolute geolocation has been corrected for both sensors, errors up to 0.5 pixels can still remain~\citep{sultan2013} and affect the selection of clean pixels near the lake shore.
\par
\textbf{Comparison with \textcolor{black}{operational} snow products}. We compare our MODIS (20 winters) and VIIRS (8 winters) \textcolor{black}{ML products} to the respective snow products: MODIS snow product (collection 6, MOD10A1), VIIRS snow product (collection 1, VNP10A1F). For the regions of interest, the VIIRS snow product has some data gaps, hence the comparison is done whenever it is available. For actual snow cover mapping, errors of 7-13\% have been reported for MODIS snow product ~\citep{snowmap}. Our findings are in line with this: for the two winters 2016-17 and 2017-18 (non-transition days only) we observe an error of 14\% w.r.t.\ our ground truth, see Fig.~\ref{fig:comparison_MSP}c.
\par 
For each lake, we first estimate the percentage of frozen pixels per day using our MODIS and VIIRS \textcolor{black}{ML products}. Since a pixel-to-pixel registration is difficult in the presence of absolute geolocation shifts and/or GSD differences, the daily percentage of frozen pixels is also computed from the snow products and the MAD is estimated for each winter. See Fig.~\ref{fig:comparison_MSP}d for the \textcolor{black}{comparison of our ML product (MODIS) and the MODIS snow product}. For the three lakes, the per-winter MAD is shown on the $y$-axis against the winters on the $x$-axis. 

\par 
Overall, the 20-year time series inter-comparison (per-lake mean and standard deviation of MAD, Fig.~\ref{fig:comparison_MSP}b) does not suggest large, systematic inconsistencies. On average, our MODIS and VIIRS \textcolor{black}{ML products} deviate by mean MAD values of 14-18\% and 12-19\% respectively. These deviations are only a little higher than the estimated error of the snow products and are relatively stable across different years.
\par
It is important to point out that the snow products are an imperfect proxy for lake ice because a lake can be frozen but not snow-covered, especially near freeze-up when it has not yet snowed onto the ice. Also, mixed ice and water cases go undetected in the MODIS snow product~\citep{snowmap}. Fig.~\ref{fig:comparison_MSP}c shows that the snow products are less consistent with the manually annotated ground truth than our ice maps. Most deviations between our estimates and the snow products occur around the transition dates, mostly freeze-up. Additionally, MODIS and VIIRS snow products use a less conservative cloud mask than we do (accepting not only \textit{confident clear} and \textit{probably clear}, but also \textit{uncertain clear} as cloud-free). Despite these issues, the inter-comparison provides a second check for our \textcolor{black}{ML products}. For completeness, we note that our algorithm has a similar issue and thin ice is sometimes confused with open water: firstly, snow-free ice is rare and underrepresented in the training set. Secondly, it appears predominantly near the transition dates (especially freeze-up) when we do not have pixel-accurate ground truth. Thirdly, thin ice and open water are difficult to distinguish, we observed that even human interpreters at times confused them when interpreting webcam images.
\par
It is interesting to note that, for both sensors, the mean MAD is inversely proportional to the lake area (see Fig.~\ref{fig:comparison_MSP}b). This hints at residual errors in the products' geolocation, which would affect smaller lakes more due to the larger fraction of pixels near the lake outline. Besides the $<0.5$-pixel inaccuracy of our maps, inaccurate geolocation of the snow products has been reported (more for MODIS, less for VIIRS) especially for freshwater bodies, due to uncertainties in gridding, reprojection etc.~\citep{snowmap}.
\par
\textbf{LIP trends using MODIS data}.
As discussed in Section \ref{sec:LIP_estimation}, we fit the "U with wings" polygon to each winter to estimate the four critical dates: FUS, FUE, BUS and BUE. Sometimes, these phenological dates are defined such that a second, consecutive day with similar ice conditions is required to confirm the event. We do not enforce this constraint, because, quite often, the days after a potential freeze-up or break-up date are cloudy, and looking further ahead runs the risk of pruning the correct candidates. 
\par
\begin{figure*}[t]
\centering
  \subfloat[Lake Sils]{\includegraphics[width=0.45\linewidth]{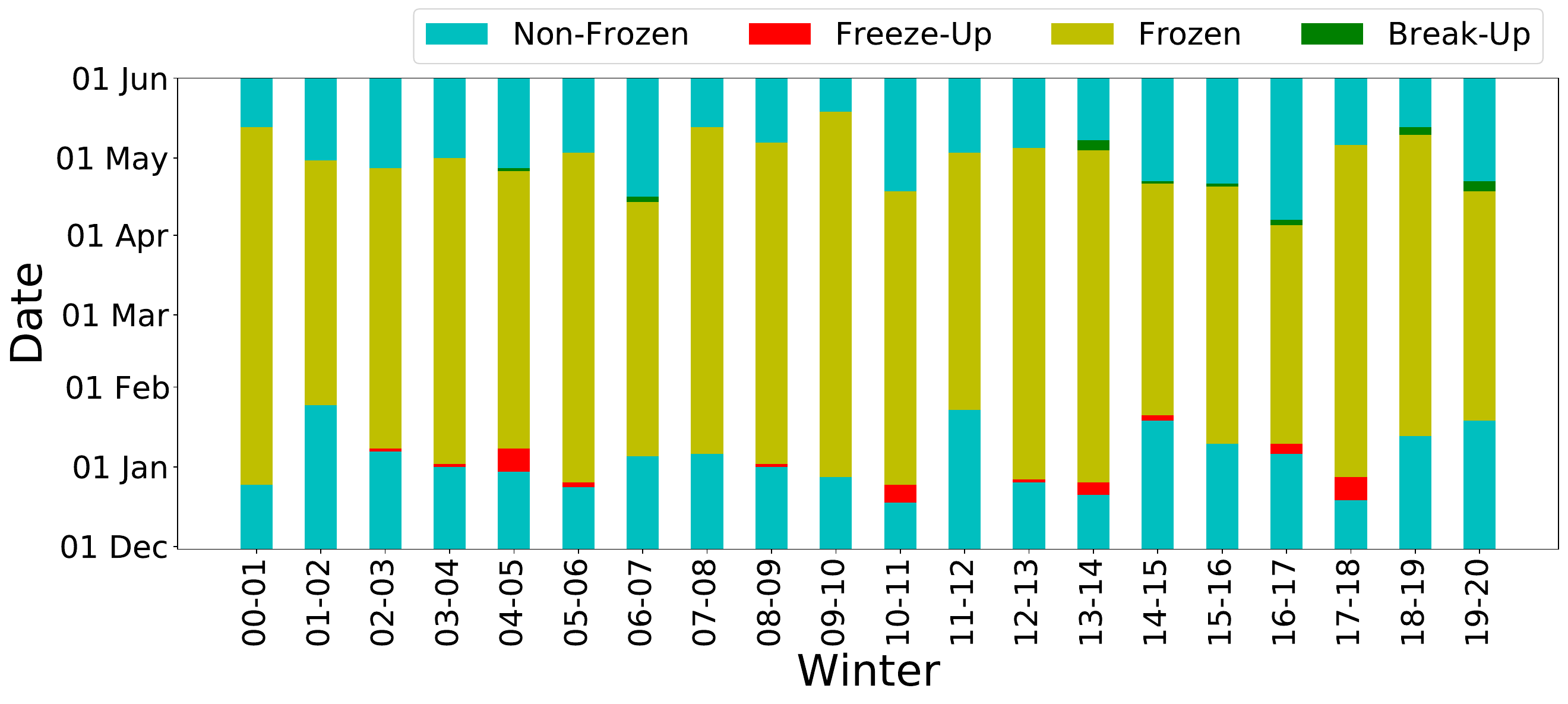}}
  \subfloat[Lake Silvaplana]{\includegraphics[width=0.45\linewidth]{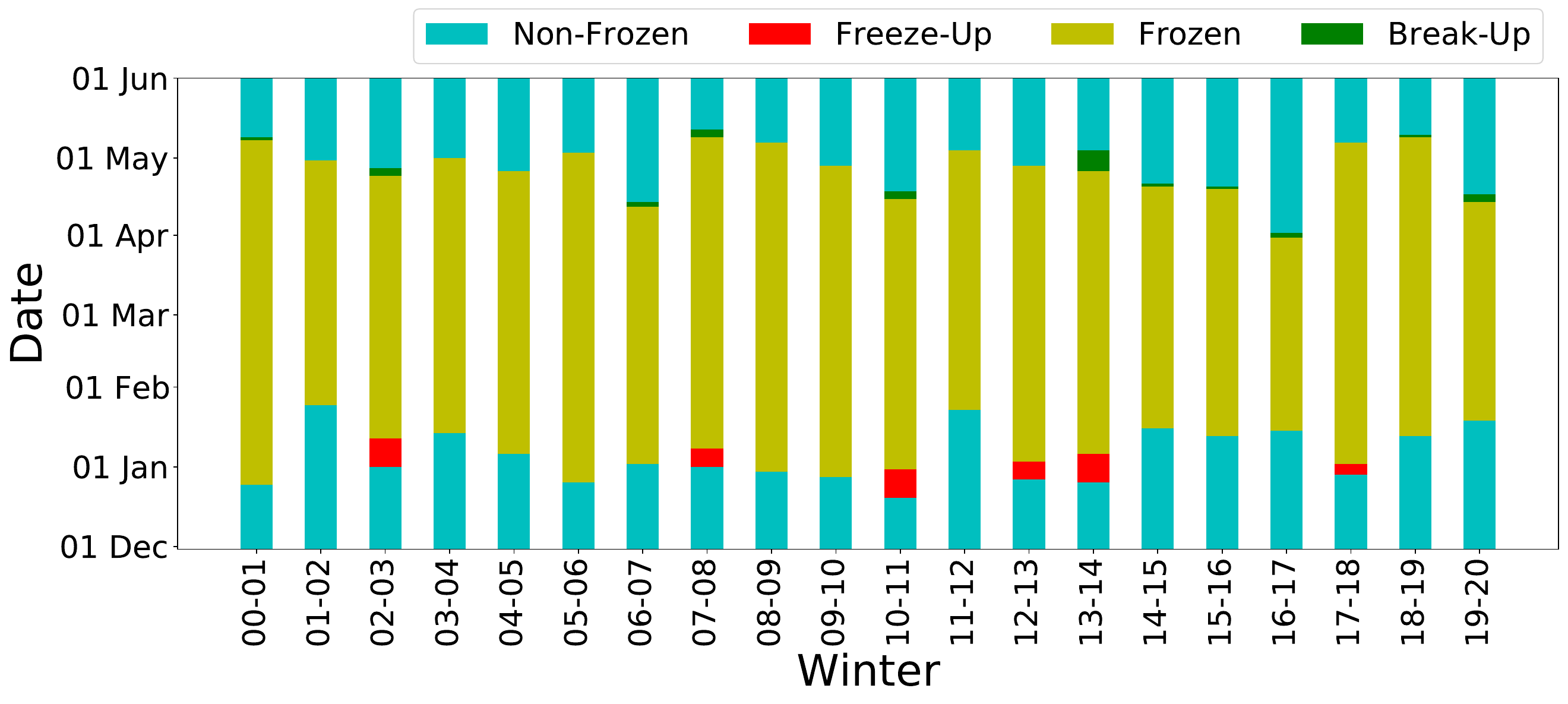}}
  \caption{\textcolor{black}{Temporal LIP characteristics estimated from MODIS using linear SVM classifier (20 winters).}}
\label{fig:trends_Sils_type2}
\end{figure*}

Using the estimated LIP dates from 20 winters, we plot their temporal evolution for lakes Sils and Silvaplana in Fig.~\ref{fig:trends_Sils_type2}. On the $y$-axis, all the dates from 1 December to 1 June (we skip September till November since no LIP events were detected during these months), while on the $x$-axis we show the winters in chronological order. In each winter, the non-frozen, freeze-up, frozen and break-up periods are displayed in blue, red, blue and dark green colours, respectively.
\par
It can be seen from Fig.~\ref{fig:trends_Sils_type2} that the freeze-thaw patterns of both lakes vary considerably across winters. For lake Sils (Silvaplana), on average, the FUS occurred on 3 January (5 January) followed by a freeze-up period of 3 (3) days until FUE on 6 January (8 January). Additionally, on average, the lake remained fully frozen (CFD) for 113 (108) days until BUS on 29 April (26 April) and the break-up period lasted 1 (1) day until BUE on 30 April (27 April). The average number of days from FUS to BUE is 117 (112).
\par
The Oberengadin region with lakes Sils and Silvaplana is a single valley (Fig.~\ref{fig:title_figure_lakes}) and hence the two have similar weather conditions. Silvaplana is relatively deeper but has a smaller area than Sils, making them comparable in terms of volume, too. So similar LIP patterns can be expected. However, the clouds above the lakes (especially on partly cloudy days), and the associated cloud mask errors, can cause small differences. In winter 2016-17, the ice-on date of the two lakes, confirmed by visual interpretation of webcams, differ by 7 (low confidence) to 10 (medium confidence) days, see also \cite{Tom2020_lakeice_RS_journal}.
\par
In most the winters, the LIP characteristics of these lakes derived using our approach are in agreement, see Fig.~\ref{fig:trends_Sils_type2}. However, there are some outliers too ($>10$ days deviation).
A notable outlier is the break-up period in winter 2009-10. For Sils (Silvaplana), BUS and BUE were both estimated as 19 May (28 April). This drift primarily happened because of a huge data gap due to clouds and cloud mask errors. During the period from 28 April till 20 May, Silvaplana had $>30$\% cloud-free MODIS acquisitions only on 28 April, 29 April, 8 May and 20 May, and the lake was detected as non-frozen on all these dates. However, Sils had MODIS acquisitions on 28 April, 29 April, 5 May and 19 May. On 5 May the lake was detected as 100\% frozen due to a false negative cloud mask, although break-up had started on the two earlier dates (75\%, respectively 60\% frozen) and the lake was ice-free on May 19.
We also checked the Landsat-7 acquisitions on 20 April 2010 and 22 May 2010 and found that both lakes were fully covered by snow on the former date and fully non-frozen on the latter date. No cloud-free Landsat-7 data is available between these two dates. For Sils, the actual BUS probably happened on 29 April ($>30$\% non-frozen) and BUE soon after (likely on 30 April, since the BUE of Silvaplana was detected as 28 April and Sils was detected $<70$\% non-frozen on 29 April). However, both dates went undetected until 19 May, because of the clouds in combination with the maximum allowed duration of 2 weeks for the break-up.
\par
In winter 2003-04, the freeze-up periods of Sils (FUS on 1 January, FUE on 2 January) and Silvaplana (FUS and FUE on 14 January) were also detected far apart, again due to a data gap because of clouds. Sils was estimated 68\% and 90\% frozen on 1 and 2 January, respectively, so they were chosen as FUS and FUE. On lake Silvaplana, the sequence for 1-5 January was 4\%$\rightarrow$13\%$\rightarrow$0\%$\rightarrow$21\%$\rightarrow$0\% frozen. Then 14 January and 21 January were both found 100\% frozen, so the fitting chose 14 January as both FUS and FUE. No cloud-free MODIS data exist on the intermediate dates 6-13 January and 15-20 January, and we could also not find any cloud-free Landsat-7 images between 21 December 2003 and 29 January 2004 (both inclusive) to check but could confirm 0\% ice cover on 20 December and 100\% cover on 30 January. Connecting all the dots, we speculate that the FUS and FUE of Silvaplana occurred soon after 5 January.
\par 
In winter 2013-14, our method asserts FUE of Sils on 26 December and of Silvaplana on 6 January. Between those dates, there were a number of partially frozen dates, but with more ice cover for Sils than Silvaplana. Additionally, 2-5 January were cloudy, leading the fitting to choose the earlier date for the former, but the later one for the latter. We again checked with Landsat-7 that on 15 December both lakes were fully non-frozen, whereas on 25 January both lakes were fully snow-covered. There exist no cloud-free Landsat-7 image in between these two dates to pin down the dates more accurately.
\par
In some winters, there is almost no freeze-up and/or break-up period detected by our algorithm. This is partly a byproduct of the relatively loose threshold needed to estimate the initial candidates for our small lakes (see Section~\ref{sec:LIP_estimation}), bringing the start and end dates of the transition closer together; and also influenced by frequent cloud cover during the critical transition dates (often more than half of all days c.f.\ Section~\ref{fig:20winter_pixelstats}c). For instance, if a couple of adjacent dates are cloudy during break-up (and the real BUS occurred during one of these dates) and on the next non-cloudy day, the lake is estimated 70.1\% non-frozen, then our fitting will choose this date as both BUS and BUE.  
\par
\begin{figure*}[t]
\centering
  \subfloat[Lake Sils]{\includegraphics[width=0.45\linewidth]{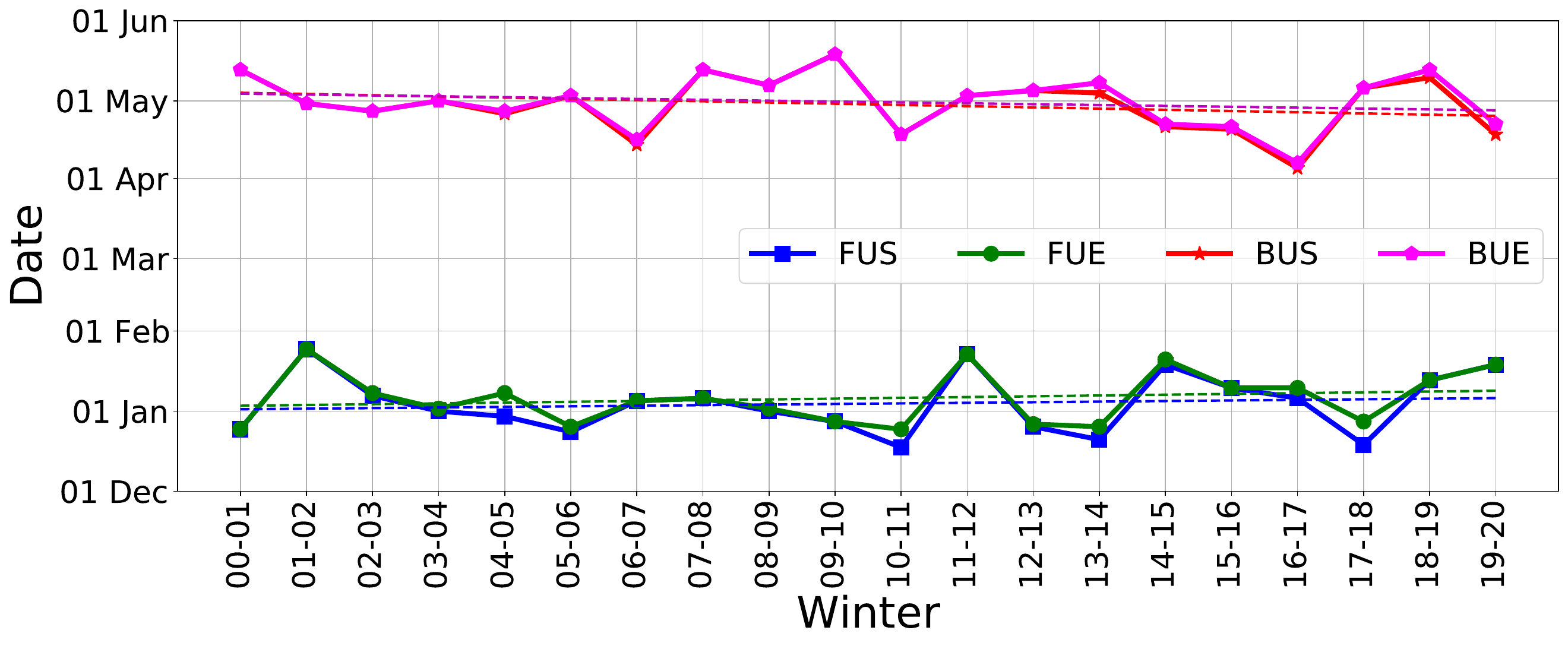}}
  \subfloat[Lake Silvaplana]{\includegraphics[width=0.45\linewidth]{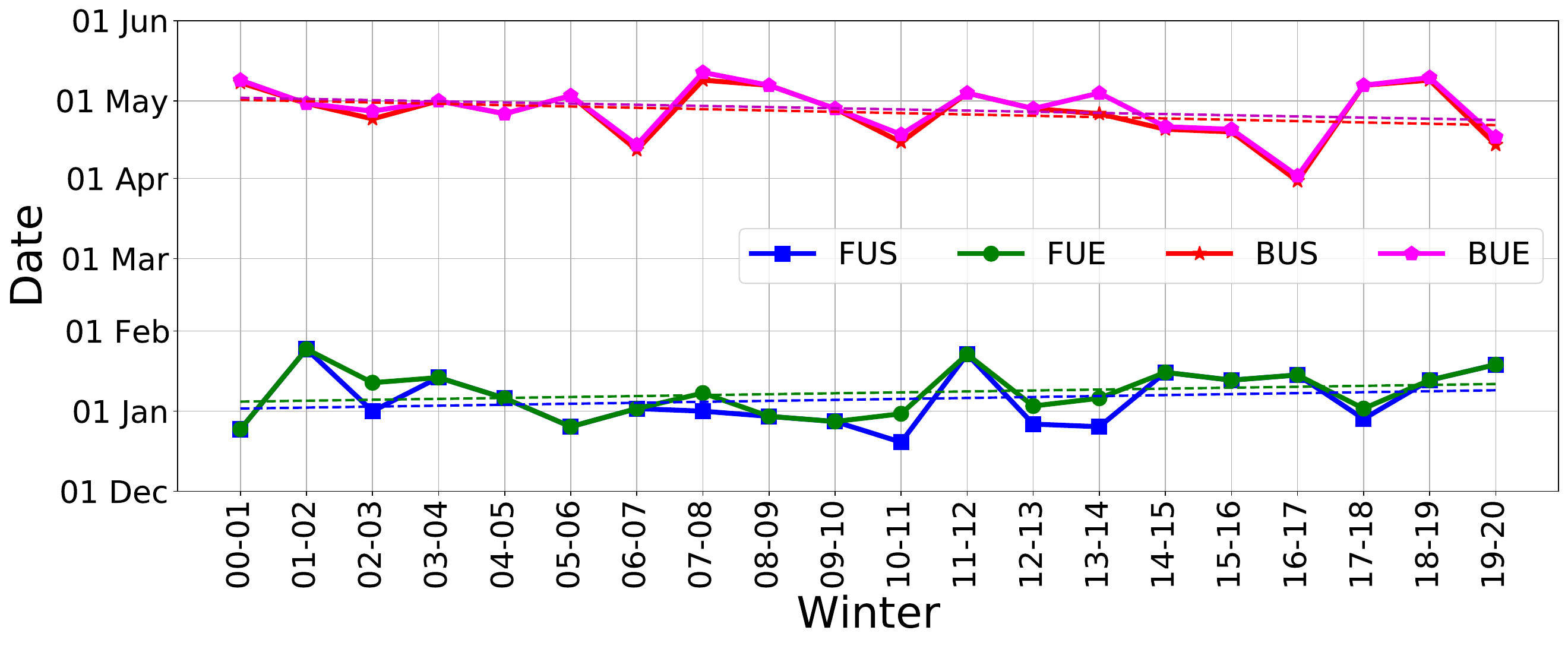}}
  \caption{\textcolor{black}{Ice freeze-up (FUS, FUE) and break-up (BUS, BUE) trends (20 winters).}}
\label{fig:trends_Sils_type1}
\end{figure*}

We go on to analyse the freeze-up and break-up patterns, by plotting the time series of the four critical dates over the past 20 winters for the same two lakes, see Fig.~\ref{fig:trends_Sils_type1}.  
Additionally, per phenological date, we fit a linear trend. Progressively later freeze-up and earlier break-up are apparent for both lakes.
\par
\begin{figure*}[t]
\centering
  \subfloat[Lake Sils]{\includegraphics[width=0.45\linewidth]{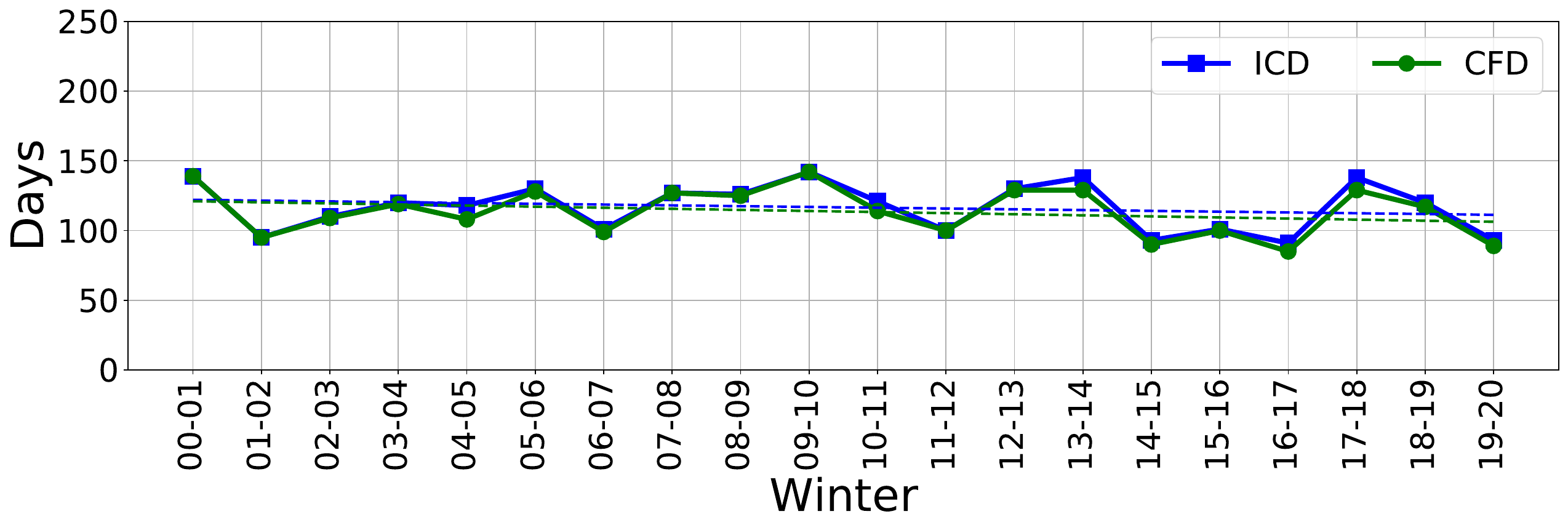}}
  \subfloat[Lake Silvaplana]{\includegraphics[width=0.45\linewidth]{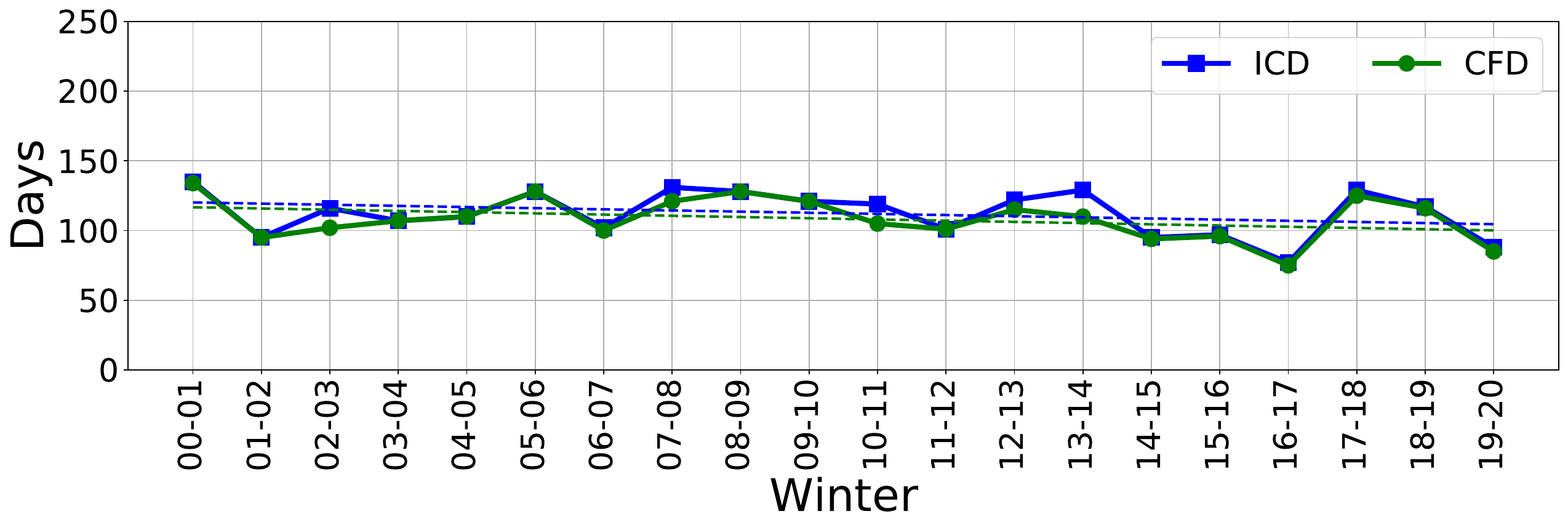}}
  \caption{\textcolor{black}{Freeze duration (ICD, CFD) trends (20 winters).}}
\label{fig:trends_Sils_FD_type1}
\end{figure*}

In each winter, we also derive the remaining LIP events (ICD, CFD) listed in Table~\ref{table:phenology}. Their trends are shown in Fig.~\ref{fig:trends_Sils_FD_type1}, with the duration in days on the $y$-axis and the winters on the $x$-axis. Obviously, ICD and CFD are decreasing for both lakes. 
\par
Quantitative trend values that we estimated for all the LIP events of Sils and Silvaplana are shown in Table~\ref{table:LIPtrends}. As explained above, we correct obvious failures of the automatic analysis, and set the following corrections for lake Sils:  BUS and BUE occurred on 29 April and 30 April respectively in winter 2009-10. Similarly, for Silvaplana: FUS and FUE occurred on 6 January in winter 2003-04 and FUE occurred on 26 December in winter 2013-14. For completeness we also fit trends without the correction -- these differ only slightly and confirm that the corrections hardly impact the overall picture. The trend towards earlier break-up is more pronounced than the one towards later freeze-up, for both Sils and Silvaplana. It is interesting to note that the decrease in freeze duration is stronger for the slightly smaller lake Silvaplana.
\par
\textbf{\textcolor{black}{Correlation: LIP events and meteorological data}}.
We have also studied the (centred and normalised) cross-correlation $\in [-1,1]$ between the LIP events (corrected version) and climate variables such as temperature, \textcolor{black}{sunshine duration}, precipitation and wind during the 20 winters. The results are shown in Fig.~\ref{fig:correlation_bargraph} for lake Sils. We do not display the results for lake Silvaplana due to space reasons. Air temperature (2m above ground) and precipitation data were collected from the nearest meteorological station SIA. However, we used the sunshine and wind measurements at station SAM, since these were not available for the complete 20 winter time span at SIA. We did not use the cloud information (number of non-cloudy pixels) from MODIS data as a measure of sunshine duration, since that would ignore the evolution throughout the day, and suffers from a non-negligible amount of cloud mask errors.
\begin{table*}[t]
\small
	    \centering
	    \begin{tabular}{ccccccc} 
		\toprule
		\textbf{Lake}&\textbf{FUS}&\textbf{FUE}&\textbf{BUS}&\textbf{BUE} & \textbf{ICD}& \textbf{CFD}\\ 
		\midrule
		Sils & 0.23 & 0.31 & $-0.46$/\textcolor{gray}{$-0.47$} & $-0.32$/\textcolor{gray}{$-0.34$} & $-0.55$/\textcolor{gray}{$-0.57$} & $-0.76$/\textcolor{gray}{$-0.78$} 
		\\ 
		Silvaplana & 0.45/\textcolor{gray}{0.37} & 0.38/\textcolor{gray}{0.36} & $-0.51$ & $-0.45$ & $-0.9$/\textcolor{gray}{$-0.82$}& $-0.89$/\textcolor{gray}{$-0.87$}\\
		\bottomrule
	    \end{tabular}
	    \caption{\textcolor{black}{Estimated LIP trends (black) and results before manual correction of the automatic results (\textcolor{gray}{grey}).}}
	    \label{table:LIPtrends}
	\end{table*}
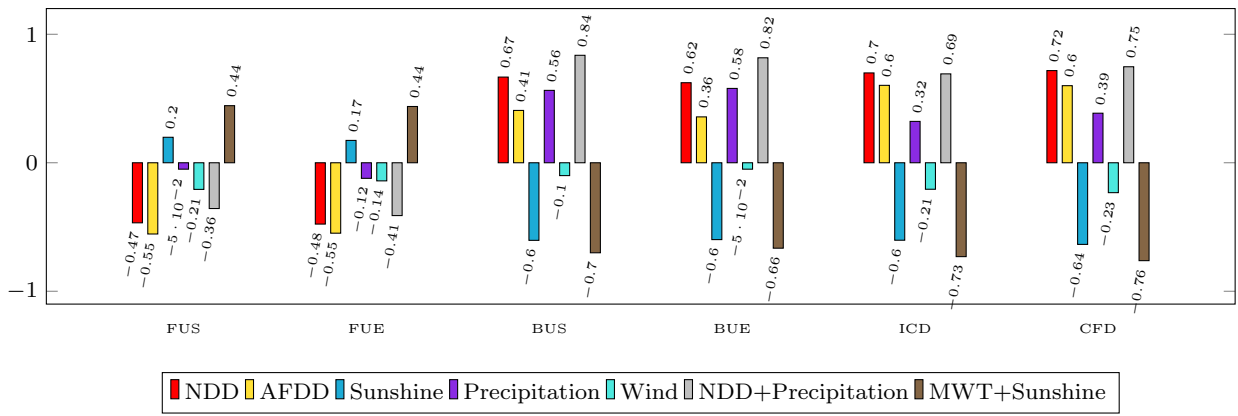
\begin{figure*}[t]
    \begin{tikzpicture}
        \centering
        \begin{axis}
        [
            width  = 0.99\linewidth,
            height = 5.5cm,
            major x tick style = transparent,
            ybar,
            bar width=3.75pt,visualization depends on=y\as\myy,
            nodes near coords,
            nodes near coords style = {rotate=80,anchor ={90+sign(\myy)*90},font = \tiny},
            ymajorgrids = false,
            ylabel = {},
            symbolic x coords={FUS, FUE, BUS, BUE, ICD, CFD}, ytick={-1.0,0,1.0},
            yticklabel style={/pgf/number format/fixed},
            xtick = data,
            scaled y ticks = false,
            x tick label style={font=\tiny,text width=2.5cm,align=center},
            enlarge x limits=0.15,
            legend style={at={(0.5,-0.230)}, anchor=north,legend columns=-1},
            ymin=-1.1,
            ymax=1.2,
            ]

           %% MWT
           %\addplot[style={BLACK,fill=RED,mark=none}]
           %    coordinates {(FUS,0.49509987) (FUE,0.50946947) (BUS,-0.54931819) (BUE,-0.49618725) (ICD,-0.64333362) (CFD,-0.66420438)}; %(FD,-0.621779)};
            
           % NDD
           \addplot[style={BLACK,fill=RED,mark=none}]
               coordinates {(FUS,-0.46745647) (FUE,-0.47732025) (BUS,0.66683413) (BUE,0.62296994) (ICD,0.69915105) (CFD,0.71733736)}; %(FD,0.69248398)};
               
           % AFDD
           \addplot[style={BLACK,fill=YELLOW,mark=none}]
            coordinates {(FUS,-0.55446756) (FUE,-0.54821868) (BUS,0.40729278) (BUE,0.35702626) (ICD,0.60250843) (CFD,0.59992355)}; %(FD,0.554626)};
           
           % Sunshine
            \addplot[style={BLACK,fill=BLUEGREEN,mark=none}]
           coordinates {(FUS,0.19833609) (FUE,0.17396719) (BUS,-0.6038375) (BUE,-0.5981061) (ICD,-0.6028947) (CFD,-0.63541562)}; %(FD,-0.489473)};
            
           % Precipitation
           \addplot[style={BLACK,fill=PURPLE,mark=none}]
            coordinates {(FUS,-0.05) (FUE,-0.12120711) (BUS,0.56338294) (BUE,0.57857537) (ICD,0.3220007) (CFD,0.38570667)}; %(FD,0.446204)};
            
            % Wind
            \addplot[style={BLACK,fill=CYAN,mark=none}]
           coordinates {(FUS,-0.2072782) (FUE,-0.14096164) (BUS,-0.10) (BUE,-0.05) (ICD,-0.20617412) (CFD,-0.23276621) };
        
            % NDD+Precipitation
            \addplot[style={BLACK,fill=SILVER,mark=none}]
           coordinates {(FUS,-0.35621634) (FUE,-0.41111539) (BUS,0.8363895) (BUE,0.81689638) (ICD,0.69174507) (CFD,0.74722025) };
        
            %  MWT+Sunshine
            \addplot[style={BLACK,fill=BLUEBROWN,mark=none}]
           coordinates {(FUS,0.44433213) (FUE,0.4379249) (BUS,-0.70067863) (BUE,-0.66491279) (ICD,-0.73064407) (CFD,-0.76194677) };
        
        \legend{NDD, AFDD, Sunshine,  Precipitation, Wind, NDD+Precipitation, MWT+Sunshine}    
        \end{axis}
    \end{tikzpicture}
    \caption{\textcolor{black}{Bar graphs showing the 20 winter correlation (y-axis) of the LIP events with climate variables (x-axis) for lake Sils. NDD and AFDD represent Negative Degree Days and Accumulated Freezing Degree Days respectively.}}
    \label{fig:correlation_bargraph}
\end{figure*}
\par
\begin{figure}[t]
\centering
  \includegraphics[width=0.49\linewidth]{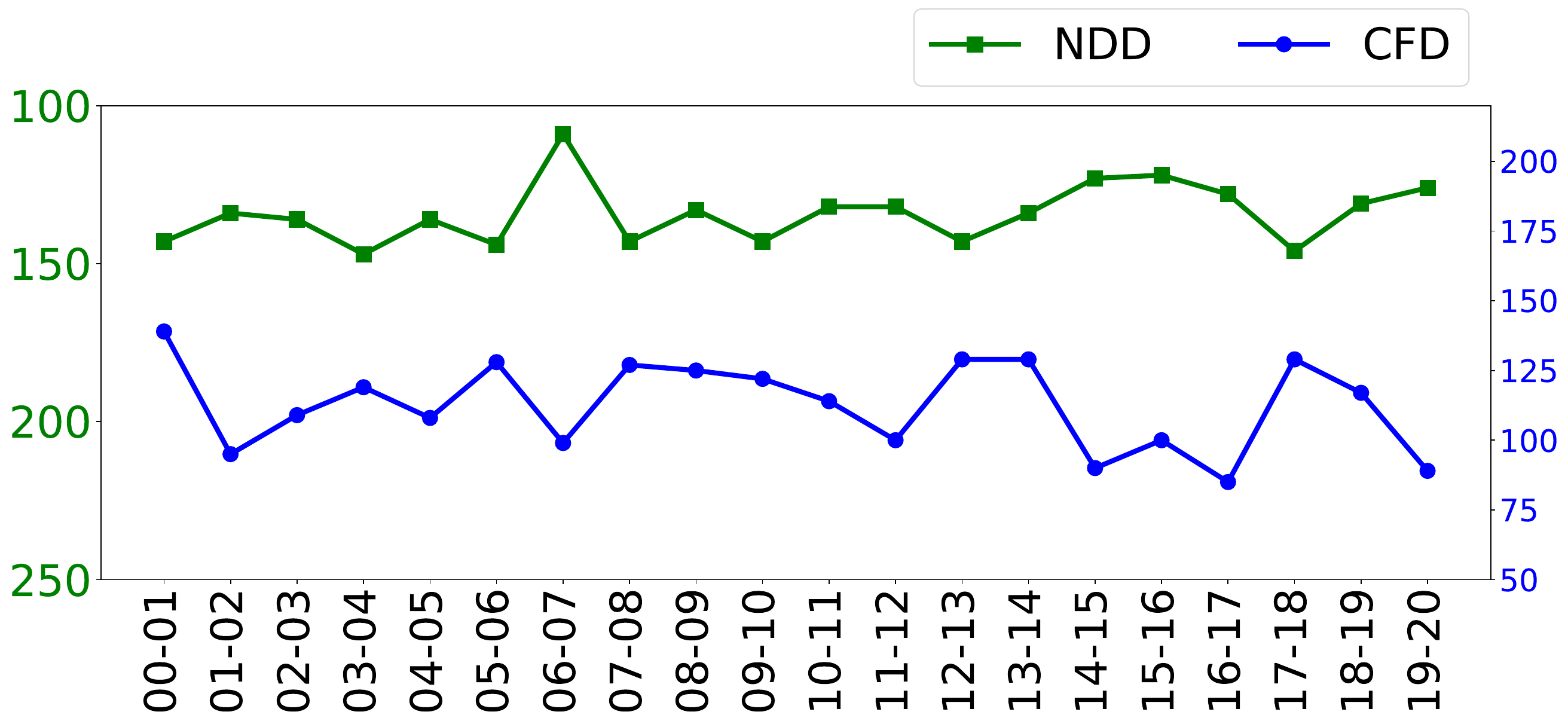}
  \includegraphics[width=0.49\linewidth]{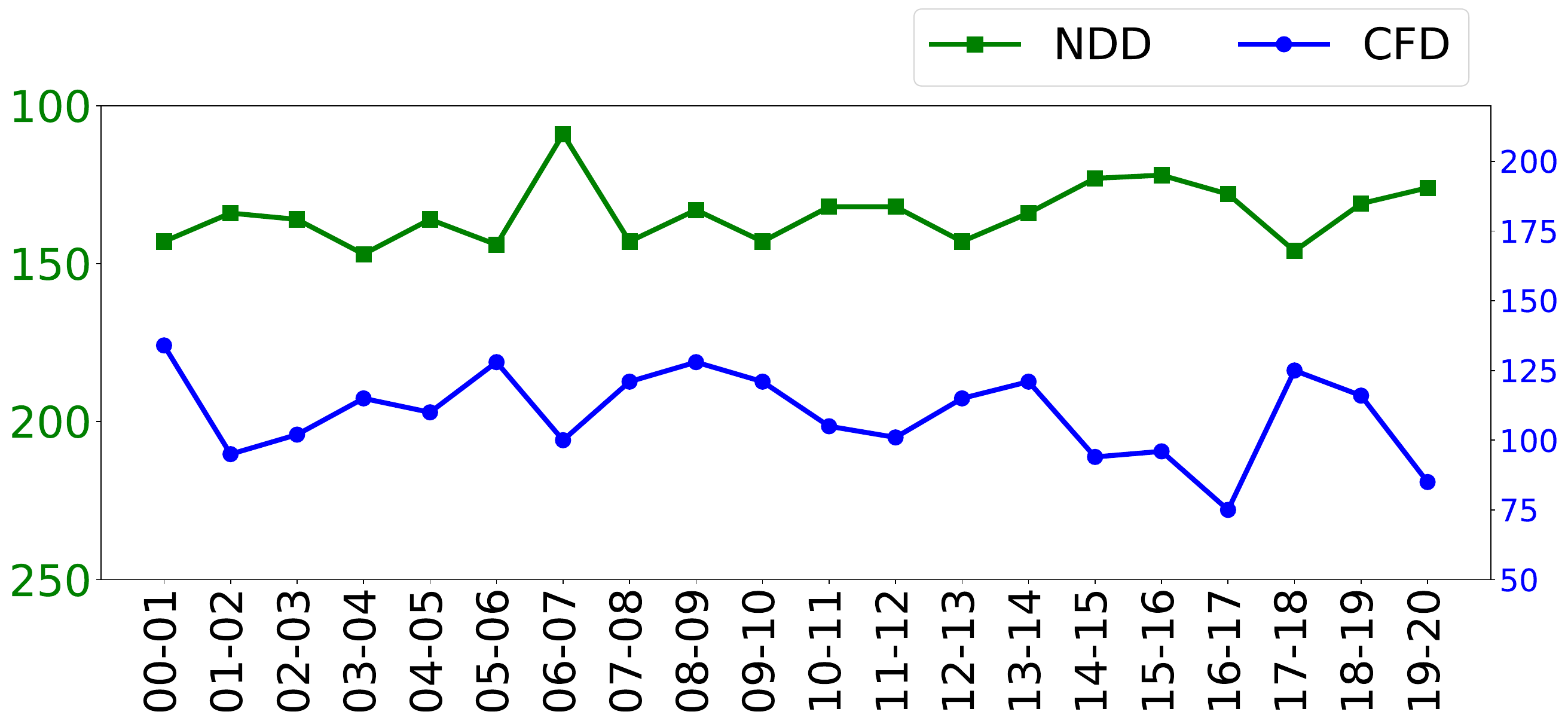}\\
  \includegraphics[width=0.49\linewidth]{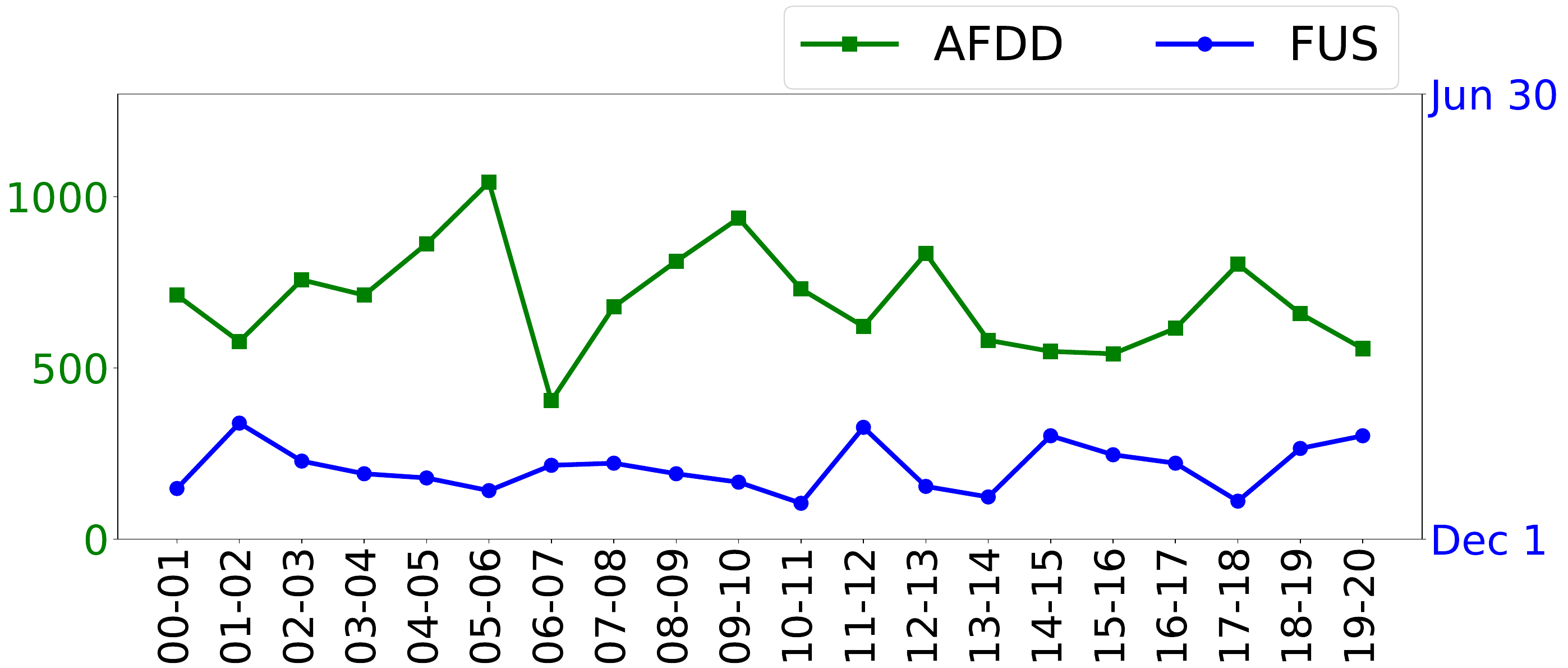}
  \includegraphics[width=0.49\linewidth]{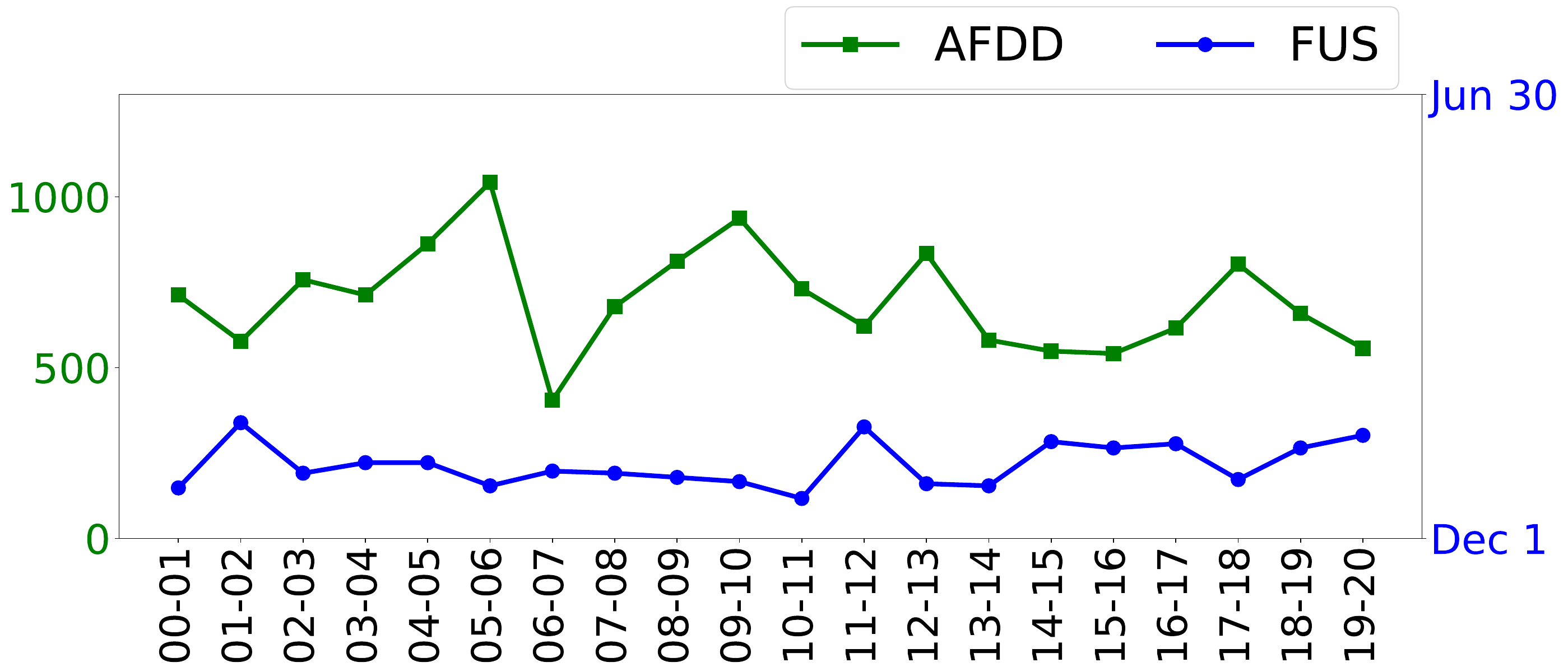}\\
  \includegraphics[width=0.49\linewidth]{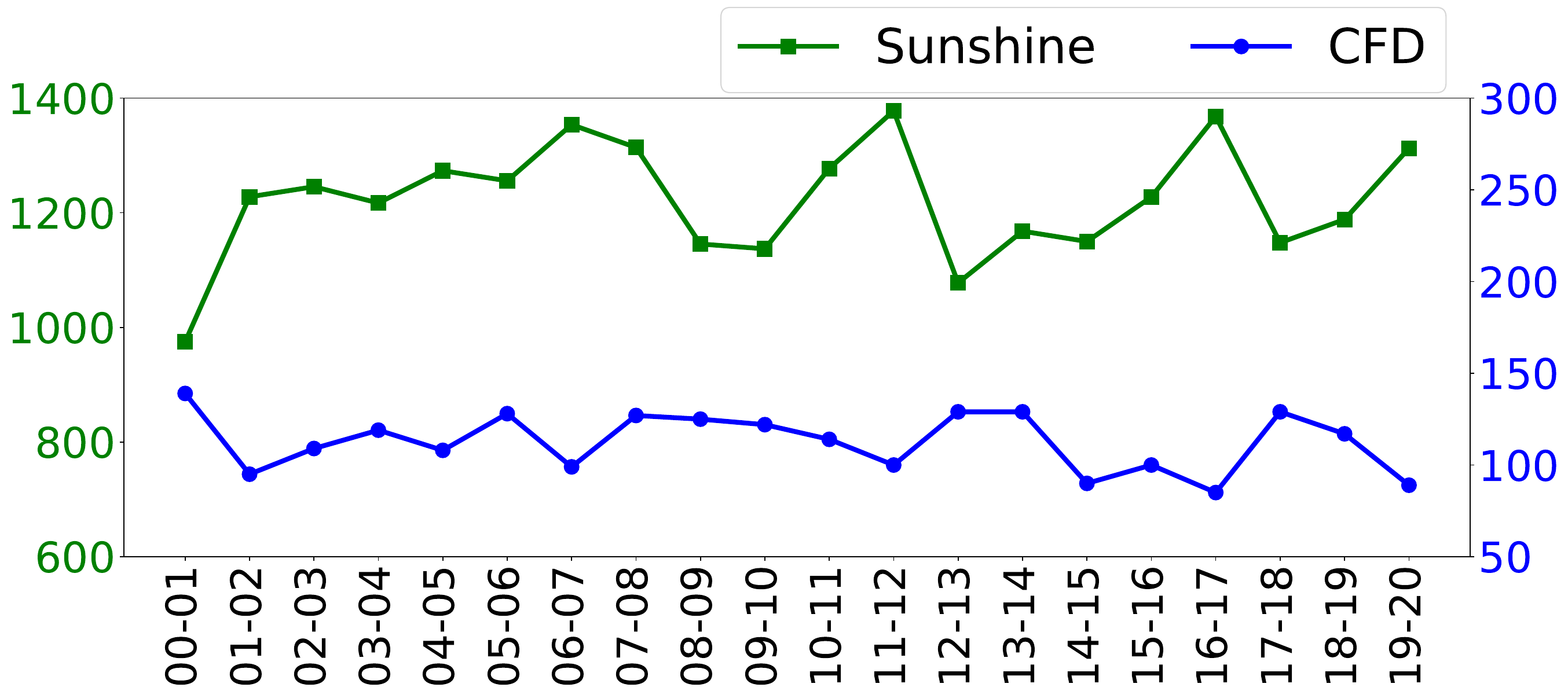}
  \includegraphics[width=0.49\linewidth]{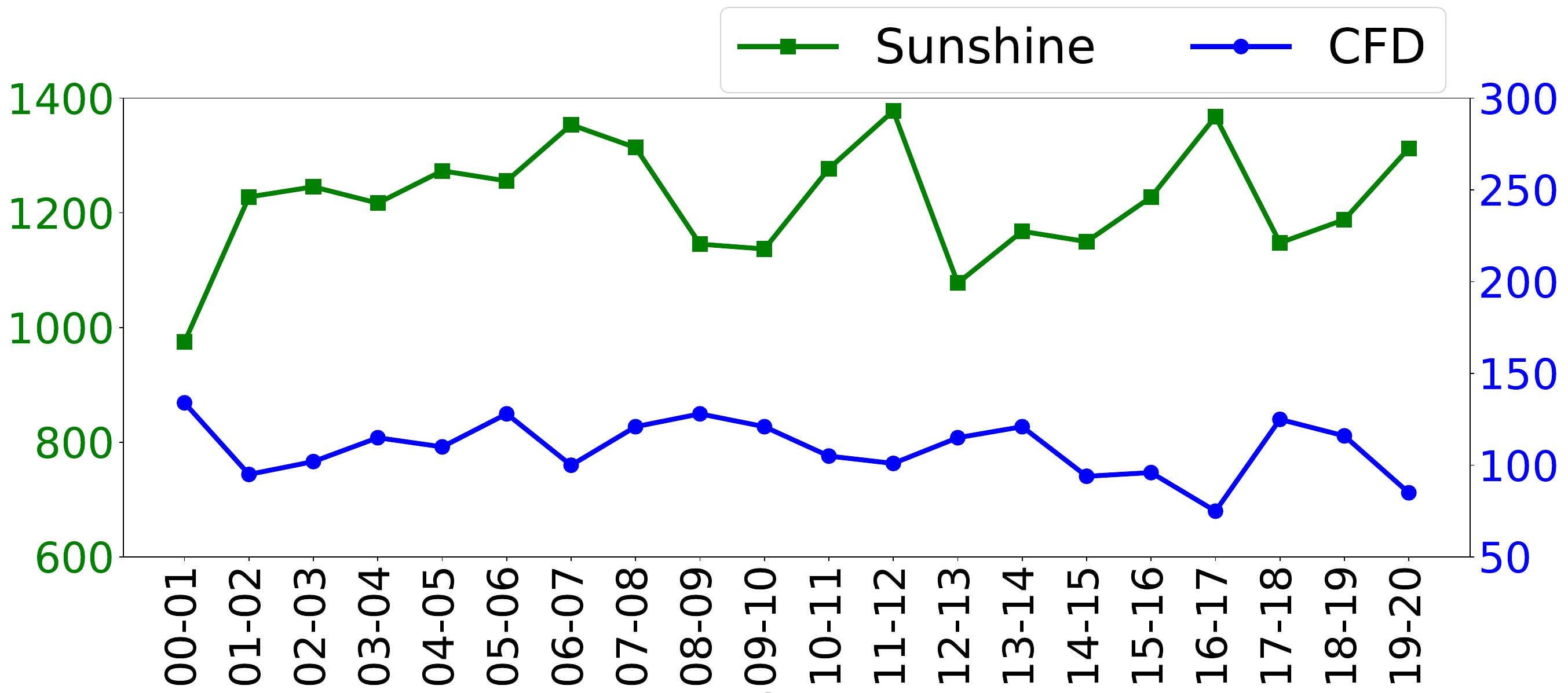}\\
  \includegraphics[width=0.49\linewidth]{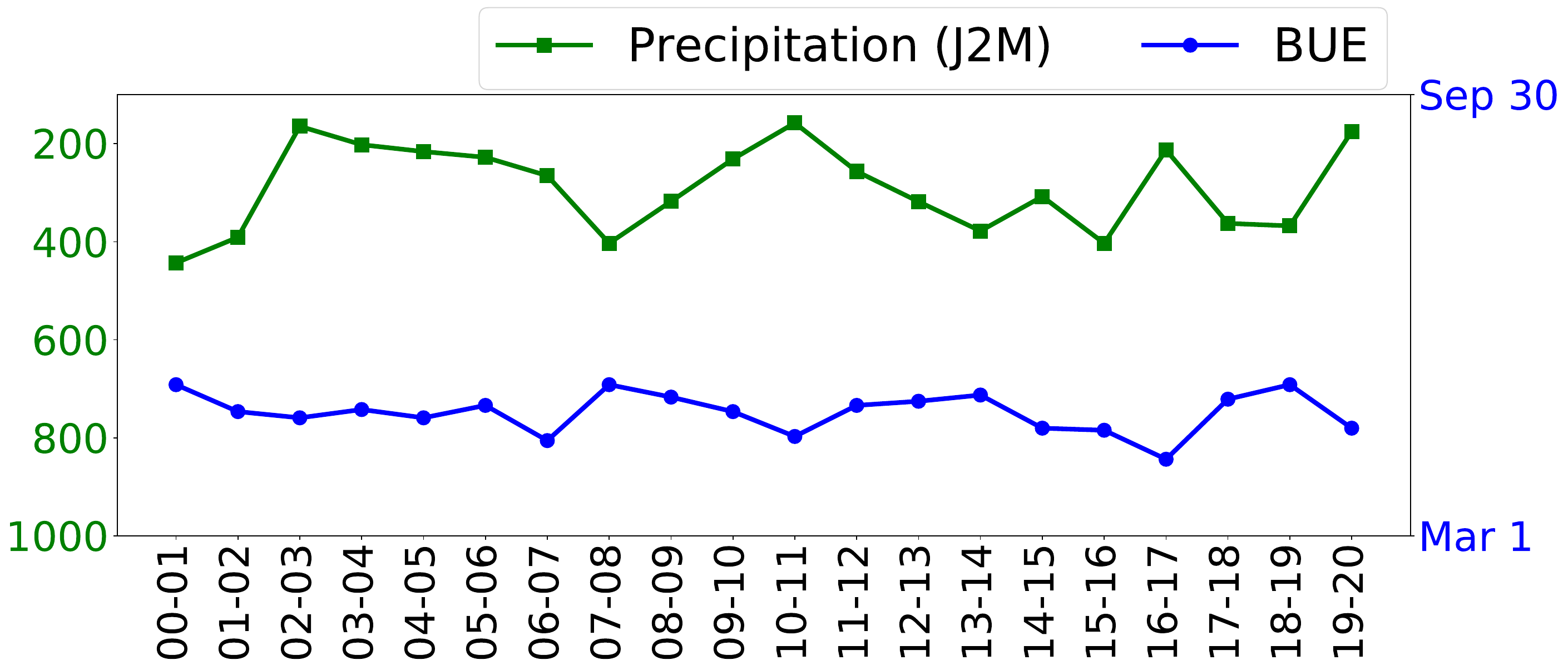}
  \includegraphics[width=0.49\linewidth]{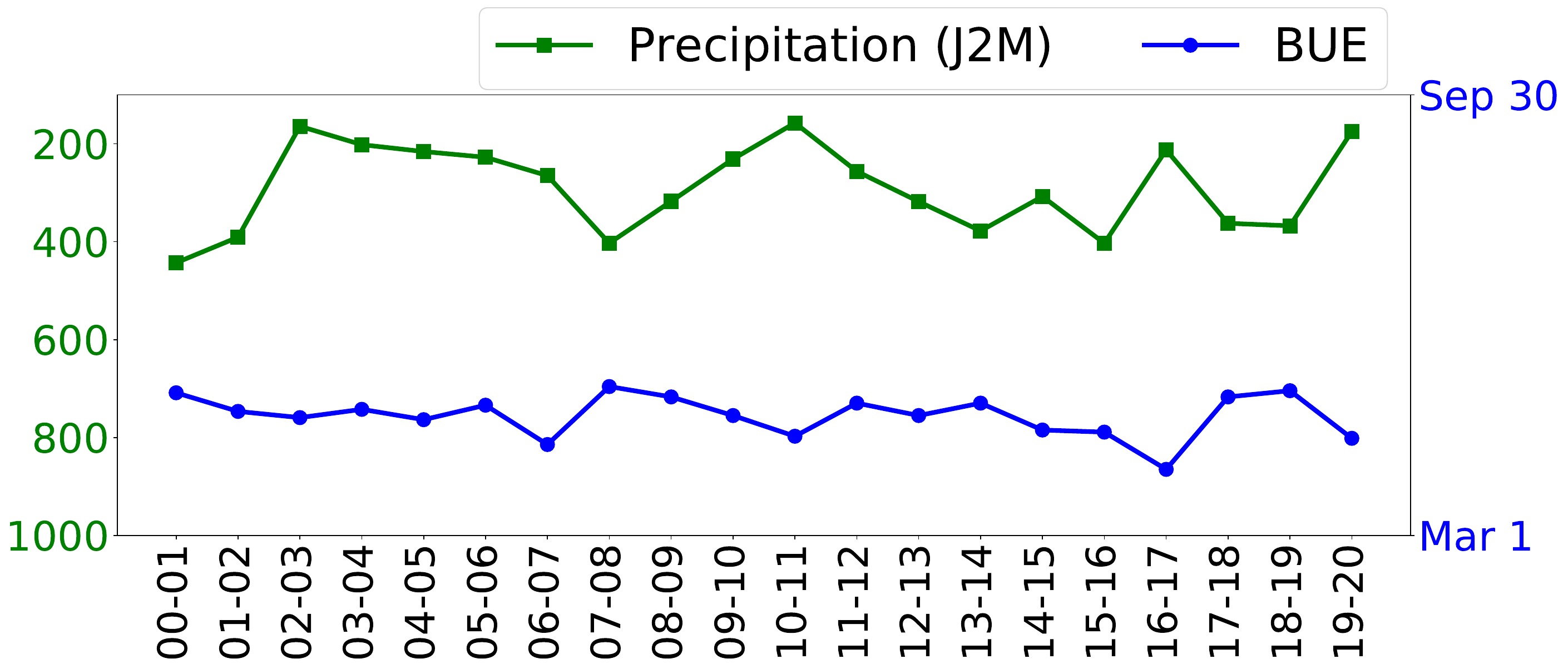}
  \caption{\textcolor{black}{Relationship between the }LIP events and weather variables: NDD (\textcolor{black}{Negative Degree Days}, days) and CFD (days) are shown in first row, AFDD (\textcolor{black}{Accumulated Freezing Degree Days}, $^{\circ}$C) and FUS in second row, total winter sunshine (hours) and CFD in third row, total precipitation (mm) in the months January to May (J2M) and BUE in last row. Results for lakes Sils and Silvaplana are displayed in left and right columns respectively. \textcolor{black}{In all the sub-figures, x-axis displays the winters from 2000-01 (00-01) till 2019-20 (19-20).}} 
\label{fig:correlation_MWT}
\end{figure}

\textcolor{black}{Negative Degree Days (NDD) corresponds to the total number of days in a winter with sub-zero air temperature (daily mean, $^{\circ}$C). As expected, Fig.~\ref{fig:correlation_bargraph} shows that NDD has a strong positive correlation with the freeze durations, and break-up events, and a negative correlation with the freeze-up events. We conclude that, indeed, as winters got warmer over the past 20 years the lakes froze later and broke up earlier. The relationship of NDD with CFD is shown in Fig.~\ref{fig:correlation_MWT}. 
}
\par
AFDD represents the cumulative sum (of daily mean temperature) on the days with average air temperature below the freezing point ($0^{\circ}$C) in a winter season. AFDD is a popular proxy for ice thickness \citep{Beyene_2018,Qinghai2020}. For both Sils and Silvaplana, AFDD has strong positive correlations with ICD and CFD, strong negative correlation with the freeze-up events, and a moderate positive correlation with ice break-up events, see Fig.~\ref{fig:correlation_bargraph}, again indicating that in colder winters (higher AFDD) the freeze-up occurs earlier and the break-up later, leading to longer freeze duration. The relatively weaker correlation for the break-up indicates that freeze-up played a larger role in that event. As an example, the \textcolor{black}{relationship} with FUS is shown Fig.~\ref{fig:correlation_MWT}. 
\par
To study the effect of sunshine on LIP events, we correlate the total winter sunshine (hours) with the freeze length events ICD and CFD, total sunshine in the months of September to December (S2D) with the freeze-up events, and the total sunshine from January to May (J2M) with the break-up events. Here, we assume that the sunshine in the months after freeze-up has no connection with freeze-up events. Similarly, we assume that the sunshine in the early winter months (September till December) does not affect the break-up events. We notice a strong negative correlation of the total winter sunshine with ICD, CFD and break-up events. The more sunshine in the months near break-up, the earlier the ice/snow melts, which also reduces the total freeze duration. An example \textcolor{black}{relationship} with CFD is visualised in Fig.~\ref{fig:correlation_MWT}. 
\par
We also check the relationship between the LIP events and total precipitation during the winter months. Similar to sunshine analysis, we correlate the total precipitation during the months from September till December, January till May and September till May to the freeze-up, break-up and freeze duration events respectively, see Fig.~\ref{fig:correlation_bargraph}. Notable are the break-up events with a good positive correlation. More precipitation in the months January to May (likely to be predominantly snow), favours later break-up, and vice-versa. \textcolor{black}{The trend for} BUE is shown in Fig.~\ref{fig:correlation_MWT}. 
\par
Inspired by \cite{Gou2015}, we also looked at the effect of wind on the LIP events, which may also influence lake freezing. We correlated the mean winter wind speed (km/h) with CFD and ICD, mean wind speed from September to December with FUE and FUS, and mean wind speed from January to May with BUS and BUE. However, we did not find any significant correlations, see Fig.~\ref{fig:correlation_bargraph}.
\par
\textcolor{black}{Finally, we explore the correlation of LIP events with the weighted combination of variables (after standardisation by mean and standard deviation) such as NDD, sunshine, precipitation and Mean Winter Temperature (MWT), see Fig.~\ref{fig:correlation_MWT} for results. MWT corresponds to the air temperature ($^{\circ}$C) averaged over the full winter. It can be noted ( Fig.~\ref{fig:correlation_MWT}) that the weighted combination of NDD and precipitation (equal weights) has a strong positive correlation with the break-up and freeze duration events indicating that during the colder winters (high NDD) with higher precipitation (probably more snow) the break-up occurs later, essentially leading to higher freeze duration. On the other hand, the weighted combination of MWT and Sunshine (equal weights) has a strong negative correlation with the break-up and freeze-duration events. Hotter winters (high MWT) with more sunshine speed up the ice break-up, thus reducing the freeze duration and vice-versa.} 
\vspace{-1.5em}
\subsection{\textcolor{black}{\textcolor{black}{Ablation studies with MODIS data from 2 winters}}}
\vspace{-0.5em}
\textcolor{black}{As a first ablation study}, we combine the data (independently for MODIS and VIIRS) of all the available lakes from winters 2016-17 and 2017-18 and perform $4$-fold cross-validation and report the overall accuracy and mIoU, see Table~\ref{table:quant_results_VIIRS_MODIS_CV}.
\par
To study how well the classifier generalise across space and time, we train a model on all except one lake~(respectively, winter) and test on the held-out lake~(winter).
Fig.~\ref{fig:MODIS_VIIRS_LOLO} displays the results (bar graphs showing overall accuracy and mIoU) of the  classifier for \textit{leave-one-lake-out} setting on MODIS (top) and VIIRS (bottom) data. It can be seen that the performance varies across lakes and sensors. 
\par
On both sensors, the best performance (especially in terms of mIoU) is achieved for the lakes Sils and Silvaplana. This is likely due to them having the most similar characteristics and imaging conditions. I.e., pixels from one of them are representative also of the other one, such that the classifier trained in one of the two generalises well to the other. Lake St.~Moritz (only for MODIS) has too few clean pixels per acquisition to draw any conclusions about \textcolor{black}{transferability}. However, we still include it in our processing to study how far lake ice monitoring with MODIS can be pushed (in terms of lake area) -- indeed, the classification is $>\!$~82.5\% correct. Lake Sihl from the region Einsiedeln is different compared to the other three lakes from the region Engadin in terms of area, weather, surrounding topography etc., c.f.\ Section~\ref{sec:method:study_area}. Hence, the performance on lake Sihl is interesting to assess geographical \textcolor{black}{transferability} over longer distances. 
\par
As a second \textcolor{black}{transferability study}, more important for our time series analysis, we check how well the trained classifier can be transferred across different winters. We train on one winter and test the model on the held-out winter (\textit{leave-one-winter-out}), see Fig.~\ref{fig:MODIS_VIIRS_LOWO}. We only have data from two consecutive winters (2016--17, 2017--18) to perform this analysis. Still, we believe that the experiment is representative for \textcolor{black}{transferability} to unseen years, since the weather conditions in different years are largely uncorrelated (c.f.\ Fig.~\ref{fig:20winter_temperature}). In particular for the two available winters, 2017--18 was markedly colder than the previous year, see Fig.~\ref{fig:20winter_temperature}.
\begin{table}[th]
\small
	    \centering
	    \begin{tabular}{ccccc} 
		\toprule
		\textbf{Sensor}    &\textbf{Feature}     &  \textbf{Method} & \textbf{Acc} & \textbf{mIoU}\\ 
		& \textbf{vector} &  &  & \\ 
		\midrule
		MODIS & 12 bands & Linear SVM & $93.4$ & $83.9$\\ 
		\midrule
		VIIRS & 5 bands & Linear SVM  & $95.1$ & $88.4$ \\ 
		\bottomrule
	    \end{tabular}
	    \caption{Four-fold cross-validation results (in percent). Overall classification accuracy (Acc) and mean intersection-over-union (mIoU) scores are shown.}
	    \normalsize
	    \label{table:quant_results_VIIRS_MODIS_CV}
	\end{table}
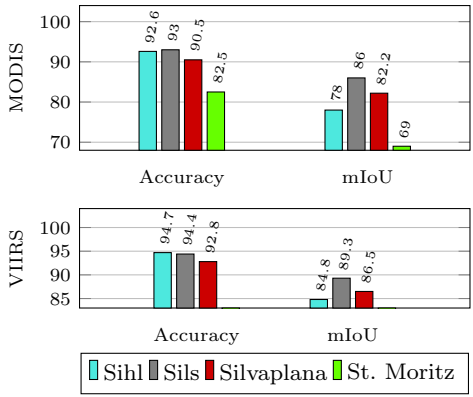
\begin{figure}[t]
\centering
  \begin{tikzpicture}
        \begin{axis}
        [
            width  = 0.8\linewidth,
            height = 3.5cm,
            major x tick style = transparent,
            ybar,
            bar width=6.5pt,
            nodes near coords,
            every node near coord/.append style={rotate=80, anchor=west,font=\tiny},
            ymajorgrids = true,
            ylabel = {MODIS},
            y label style={font=\scriptsize},
            y tick label style={font=\scriptsize,align=center},
            symbolic x coords={Accuracy, mIoU},
            xtick = data,
            scaled y ticks = false,
            x tick label style={font=\scriptsize,text width=1.8cm,align=center},
            enlarge x limits=0.55,
            legend style={at={(0.5,-0.450)}, anchor=north,legend columns=-1},
            ymin=68,
            ymax=104,
            %legend style={ area legend, at={(0.5,-0.15)}, anchor=north, legend columns=1}]
            %legend pos=outer north east,
        ]
            \addplot[style={BLACK,fill=CYAN,mark=none}]
                coordinates {(Accuracy,92.6) (mIoU,78.0)};%  10
            \addplot[style={BLACK,fill=DARKSILVER,mark=none}]
                coordinates {(Accuracy,93) (mIoU,86)};%  11
            \addplot[style={BLACK,fill=LIGHTRED,mark=none}]
                coordinates {(Accuracy,90.5) (mIoU,82.2)};%  9
            \addplot[style={BLACK,fill=LIGHTGREEN,mark=none}]
                coordinates {(Accuracy,82.5) (mIoU,69)};%  12
            %\legend{SL, SR, RF, XG}
        \end{axis}
    \end{tikzpicture}\\
        \begin{tikzpicture}
        \begin{axis}
        [
            width  = 0.8\linewidth,
            height = 2.9cm,
            major x tick style = transparent,
            ybar,
            bar width=6.5pt,
            nodes near coords,
            every node near coord/.append style={rotate=80, anchor=west,font=\tiny},
            ymajorgrids = true,
            ylabel = {VIIRS},
            y label style={font=\scriptsize,align=center},
            y tick label style={font=\scriptsize,align=center},
            symbolic x coords={Accuracy, mIoU},
            xtick = data,
            scaled y ticks = false,
            x tick label style={font=\scriptsize,text width=1.8cm,align=center},
            enlarge x limits=0.75,
            legend style={at={(0.5,-0.450)}, anchor=north,legend columns=-1},
            ymin=83,
            ymax=104,
        ]
            \addplot[style={BLACK,fill=CYAN,mark=none}]
                coordinates {(Accuracy,94.7) (mIoU,84.8)};%  10
            \addplot[style={BLACK,fill=DARKSILVER,mark=none}]
                coordinates {(Accuracy,94.4) (mIoU,89.3)};%  11
            \addplot[style={BLACK,fill=LIGHTRED,mark=none}]
                coordinates {(Accuracy,92.8) (mIoU,86.5)};%  9
            \addplot[style={BLACK,fill=LIGHTGREEN,mark=none}]
                coordinates {(Accuracy,0) (mIoU,0)};%  9
            \legend{Sihl, Sils, Silvaplana, St.~Moritz}
        \end{axis}
    \end{tikzpicture}
    \caption{\textcolor{black}{\textcolor{black}{Transferability study (across lakes) results}.}}
	\label{fig:MODIS_VIIRS_LOLO}
\end{figure}%
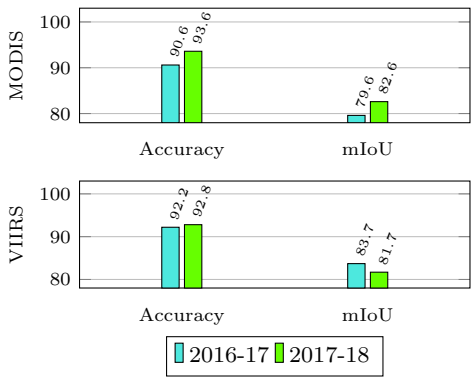
\begin{figure}[t]
\centering
  \begin{tikzpicture}
        \begin{axis}
        [
            width  = 0.8\linewidth,
            height = 3.1cm,
            major x tick style = transparent,
            ybar,
            bar width=6.5pt,
            nodes near coords,
            every node near coord/.append style={rotate=70, anchor=west,font=\tiny},
            ymajorgrids = true,
            ylabel = {MODIS},
            y label style={font=\scriptsize,align=center},
            y tick label style={font=\scriptsize,align=center},
            symbolic x coords={Accuracy, mIoU},
            xtick = data,
            scaled y ticks = false,
            x tick label style={font=\scriptsize,text width=1.8cm,align=center},
            enlarge x limits=0.55,
            legend style={at={(0.5,-0.450)}, anchor=north,legend columns=-1},
            ymin=78,
            ymax=103,
            %legend style={ area legend, at={(0.5,-0.15)}, anchor=north, legend columns=1}]
            %legend pos=outer north east,
        ]
            \addplot[style={BLACK,fill=CYAN,mark=none}]
                coordinates {(Accuracy,90.6) (mIoU,79.6)};
            \addplot[style={BLACK,fill=LIGHTGREEN,mark=none}]
                coordinates {(Accuracy,93.6) (mIoU,82.6)};
            %\legend{SL, SR, RF, XG}
        \end{axis}
    \end{tikzpicture}
    \\
    \begin{tikzpicture}
        \begin{axis}
        [
            width  = 0.8\linewidth,
            height = 3.0cm,
            major x tick style = transparent,
            ybar,
            bar width=6.5pt,
            nodes near coords,
            every node near coord/.append style={rotate=70, anchor=west,font=\tiny},
            ymajorgrids = true,
            ylabel = {VIIRS},
            y label style={font=\scriptsize,align=center},
            y tick label style={font=\scriptsize,align=center},
            symbolic x coords={Accuracy, mIoU},
            xtick = data,
            scaled y ticks = false,
            x tick label style={font=\scriptsize,text width=1.8cm,align=center},
            enlarge x limits=0.55,
            legend style={at={(0.5,-0.450)}, anchor=north,legend columns=-1},
            ymin=78,
            ymax=103,
            %legend style={ area legend, at={(0.5,-0.15)}, anchor=north, legend columns=1}]
            %legend pos=outer north east,
        ]
            \addplot[style={BLACK,fill=CYAN,mark=none}]
                coordinates {(Accuracy,92.2) (mIoU,83.7)};
            \addplot[style={BLACK,fill=LIGHTGREEN,mark=none}]
                coordinates {(Accuracy,92.8) (mIoU,81.7)};
            \legend{$2016$-$17$, $2017$-$18$}
        \end{axis}
    \end{tikzpicture}
    \caption{\textcolor{black}{Transferability study (across winters) results.}}
	\label{fig:MODIS_VIIRS_LOWO}
\end{figure}%
\par
\textcolor{black}{The classifier exhibits a certain performance drop when having to generalise beyond the exact training conditions.} Table~\ref{table:quant_results_gen_drop} shows the detailed performance drops compared to Table~\ref{table:quant_results_VIIRS_MODIS_CV}. 
\textcolor{black}{Note that in the ablation studies we must hold out some ground truth for evaluation and therefore have a smaller training set.}
\vspace{-1.5em}
\subsection{Discussion}
\vspace{-0.25em}
In any ML-based system, the variety in the training dataset has a critical influence on the model being learnt. \textcolor{black}{Our dataset consists of small lakes and has a significant class imbalance since we include all cloud-free dates from September till May, of which only a minority is frozen}. This is a biased, but realistic scenario, representative of mountain lakes in sub-Arctic and temperate climate zones. 

\begin{table}[th]
\small
	    \centering
	    \begin{tabular}{ccc} 
		\toprule
		\textbf{Sensor}    &\textbf{Loss type}    &\textbf{Linear SVM}\\ 
		\midrule
		MODIS   & across lakes & 3.7/\textcolor{gray}{5.1}\\ 
		MODIS & across winters & 1.3/\textcolor{gray}{2.8}\\
		\midrule
		VIIRS   & across lakes & 1.1/\textcolor{gray}{1.5}\\ 
		VIIRS   & across winters & 2.6/\textcolor{gray}{5.7}\\
		\bottomrule
	    \end{tabular}
	    \caption{\textcolor{black}{Transferability} loss encountered by \textcolor{black}{our} classifier. Drop (in percent) for overall accuracy and mIoU are shown in black and grey respectively.}
	    \normalsize
	    \label{table:quant_results_gen_drop}
	\end{table}
\par
Our MODIS and VIIRS \textcolor{black}{products} validate each other in a relative sense ($<5.6~\%$ MAD in the worst case, lake Sils), but could be subject to a common bias. In the absence of ground truth, there is no way to assess our absolute accuracy, but as an external check against a methodologically different mapping scheme we inter-compared our \textcolor{black}{ML products} against the respective operational snow products. The deviations were in the expected range (MAD $<20~\%$).
\par
\textcolor{black}{Some errors exist in both MODIS and VIIRS cloud masks. The most critical ones are false negatives, where an actually cloudy pixel goes undetected. Such cases can corrupt model learning and inference and introduce errors in the predicted ice maps.} The trade-off between spatial and temporal resolution makes it difficult to monitor smaller lakes -- with 21 MODIS (9 VIIRS pixels) for lake Silvaplana and only 4 MODIS pixels for lake St.~Moritz, our study goes to the limit in that respect. A further, often-named obstacle for optical satellite observation is occlusions due to clouds, which significantly reduce the effective temporal resolution and also cause irregular gaps in the time series. These unpredictable data gaps are particularly troublesome for ice phenology because the critical events occur over a short time and at times of the year \textcolor{black}{when clouds are} frequent in sub-Arctic and mid-latitudes.
\textcolor{black}{Such gaps} are the main source of error in our LIP estimation, besides cloud mask errors, confusion between open water and thin/floating snow-free-ice, and quantisation effects around hard thresholds. This makes phenological observations challenging -- in particular, the uncertainties of our predictions are largest during freeze-up, because of the frequent, but short-lived presence of snow-free ice. Still, it appears that our classifier copes better with the \textcolor{black}{ice reflectance} than simple index-based snow products. \textcolor{black}{ML is a powerful tool to recognise the underlying patterns where mechanistic models are lacking or too complicated. Furthermore, unlike the traditional threshold-based approaches which needed separate tuning for each lake and winter, our ML methodology relies on a fixed set of parameters applicable for all our target lakes and winters.} 
\par
For small lakes, even small geolocalisation errors have a large effect. Our work is also on the challenging ends of the spectrum in terms of local weather conditions: in a drier climate the observations would be less affected by clouds (we process lakes with as little as 30\% cloud-free area to obtain sufficient temporal coverage), and fewer clouds mean fewer cloud-mask errors.%

\vspace{-1.5em}
\section{Conclusion}
\vspace{-0.25em}
In this paper, we reported results for selected lakes in Southeastern Switzerland, where we have retrieved lake ice phenology based on MODIS optical image time series. On the one hand, we have \textcolor{black}{shown that, even for small high-Alpine lakes, ice cover can be derived from low spatial resolution MODIS data; and that lake ice phenology retrieved in this manner over 20 years exhibits meaningful correlations with climate data.} On the other hand, we have confirmed that a dedicated machine learning scheme maps lake ice more accurately than the classical index- and threshold-based approaches. 
\par
As expected, our results point towards later freeze-up (\textit{freeze-up start} at a rate of $0.23$ d/a for lake Sils, respectively $0.45$ d/a for Silvaplana and \textit{freeze-up end} at a rate of $0.31$ d/a for lake Sils, respectively $0.38$ d/a for Silvaplana), earlier break-up (\textit{break-up start}: $-0.46$ d/a for lake Sils, respectively $-0.51$ d/a for Silvaplana and \textit{break-up end}: $-0.32$ d/a for lake Sils, respectively $-0.45$ d/a for Silvaplana) and decreasing freeze duration (\textit{ice coverage duration}: $-0.55$ d/a for lake Sils, respectively $-0.90$ d/a for Silvaplana and \textit{complete freeze duration}: $-0.76$ d/a for lake Sils, respectively $-0.89$ d/a for Silvaplana). We also observed significant (but not surprising) correlations with climate indicators such as temperature, sunshine and precipitation. %
\par
Our approach is generic and easy to apply to other sensors beyond MODIS and VIIRS (given training data). Importantly, the VIIRS sensor is projected to ensure continuity well into the future, opening up the possibility to establish an even longer time series. One solution for the cloud issues of optical satellites is to complement/replace them with radar observations, e.g., Sentinel-1 SAR. We have done preliminary research in this direction~\citep{tom_aguilar_2020}. SAR-optical data fusion holds great promise, particularly in view of the GCOS requirement to monitor lake ice at daily temporal resolution.
%\par
We expect that machine learning-based ice detection itself could be further improved with pixel-accurate annotations during transition dates, as well as for more winters and a wider variety of lakes. Unfortunately, gathering such data is not only a considerable, tedious effort, but also poses its own challenges. In most locations and for older data, no corresponding webcam data (or similar regular photography) is available; even when available, its coverage is almost invariably incomplete; and even with usable webcam and satellite imagery, manual annotation is not trivial and prone to mistakes exactly in the situations that are most critical also for computational analysis (such as thin, black ice). We speculate that, given the enormous archive of unlabelled satellite data, approaches such as unsupervised, semi-supervised or active learning may be applicable and could improve the lake ice detector.
\vspace{-0.75em}
\begin{acknowledgements}
We are grateful to Damien Bouffard (EAWAG, Switzerland) for providing advice regarding the correlation between meteorological variables and LIP events.
\end{acknowledgements}
\vspace{-1.25em}
\section*{Declarations}
\vspace{-0.25em}
\textbf{Funding}
This work is part of the project \textit{Integrated lake ice monitoring and generation of sustainable, reliable, long time series} funded by the Swiss Federal Office of Meteorology and Climatology MeteoSwiss in the framework of GCOS Switzerland.
\vspace{0.6em}
\newline
\textbf{Conflict of interest}
The authors declare no conflict of interest.%
\vspace{-1.0em}
% BibTeX users please use one of
\bibliographystyle{spbasic}      % basic style, author-year citations
\bibliography{mybibfile}   % name your BibTeX data base

\end{document}